\documentclass[12pt,twoside]{report}

\usepackage{cvpr}
\usepackage{times}
\usepackage{epsfig}

\usepackage[export]{adjustbox}

\usepackage[pagebackref=true,breaklinks=true,letterpaper=true,colorlinks,bookmarks=false]{hyperref}

\cvprfinalcopy 

\def\cvprPaperID{****} 

\ifcvprfinal\fi

\usepackage[table]{xcolor}
 \usepackage{amssymb}
\usepackage{hhline}

\newcommand{\reporttitle}{Interpretable Deep Neural Networks for Dimensional and Categorical Emotion Recognition in-the-wild}
\newcommand{\reportauthor}{Yicheng Xia}
\newcommand{\supervisor}{Dimitrios Kollias}
\newcommand{\degreetype}{Computing Science}

\usepackage[a4paper,hmargin=2.8cm,vmargin=2.0cm,includeheadfoot]{geometry}
\usepackage{textpos}
\usepackage{tabularx,longtable,multirow,subfigure,caption}
\usepackage{fncylab} 
\usepackage{fancyhdr} 
\usepackage{url} 
\usepackage[english]{babel}
\usepackage{amsmath}
\usepackage{amssymb}
\usepackage{graphicx}
\usepackage{dsfont}
\usepackage{epstopdf} 
\usepackage{array}
\usepackage{latexsym}

\usepackage{subfiles}

\usepackage[all]{hypcap}

\usepackage{multicol}
\usepackage{booktabs}
\usepackage{listings}

\def\@makechapterhead#1{%
  \vspace*{10\p@}%
  {\parindent \z@ \raggedright \sffamily
    \interlinepenalty\@M
    \Huge\bfseries \thechapter \space\space #1\par\nobreak
    \vskip 30\p@
  }}

\def\@makeschapterhead#1{%
  \vspace*{10\p@}%
  {\parindent \z@ \raggedright
    \sffamily
    \interlinepenalty\@M
    \Huge \bfseries  #1\par\nobreak
    \vskip 30\p@
  }}

\allowdisplaybreaks

\date{September 2018}

\begin{document}

\begin{titlepage}

\newcommand{\HRule}{\rule{\linewidth}{0.5mm}} 


\includegraphics[width = 4cm]{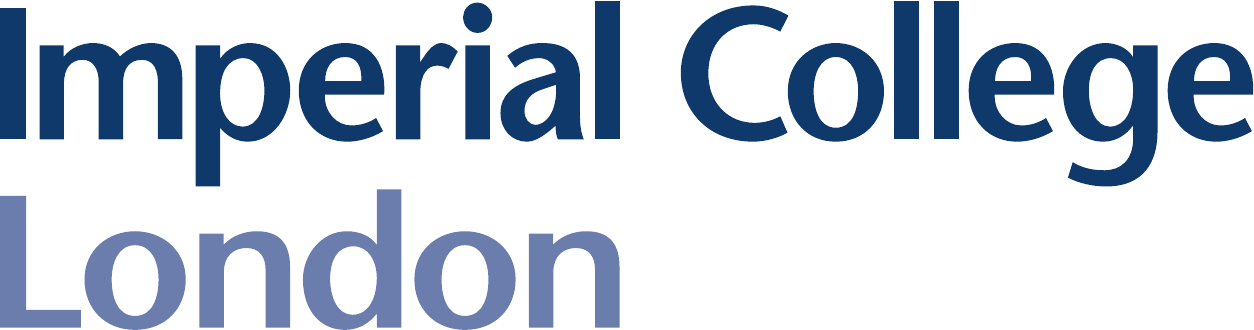}\\[0.5cm] 

\center 


\textsc{\Large Imperial College London}\\[0.5cm] 
\textsc{\large Department of Computing}\\[0.5cm] 


\HRule \\[0.4cm]
{ \huge \bfseries \reporttitle}\\ 
\HRule \\[1.5cm]
 

\begin{minipage}{0.4\textwidth}
\begin{flushleft} \large
\emph{Author:}\\
\reportauthor 
\end{flushleft}
\end{minipage}
~
\begin{minipage}{0.4\textwidth}
\begin{flushright} \large
\emph{Supervisor:} \\
\supervisor 
\end{flushright}
\end{minipage}\\[4cm]

\vfill 
Submitted in partial fulfillment of the requirements for the MSc degree in
\degreetype~of Imperial College London\\[0.5cm]

\makeatletter
\@date 
\makeatother

\end{titlepage}

\pagenumbering{roman}
\clearpage{\pagestyle{empty}\cleardoublepage}
\setcounter{page}{1}
\pagestyle{fancy}

\begin{abstract}
Emotions play an important role in people's life. Understanding and recognising is not only important for interpersonal communication, but also has promising applications in Human-Computer Interaction, automobile safety and medical research. \par

This project focuses on extending the emotion recognition database, and training the CNN + RNN emotion recognition neural networks with emotion category representation and valence \& arousal representation. The combined models are constructed by training the two representations simultaneously. The comparison and analysis between the three types of model are discussed. The inner-relationship between two emotion representations and the interpretability of the neural networks are investigated. The findings suggest that categorical emotion recognition performance can benefit from training with a combined model. And the mapping of emotion category and valence \& arousal values can explain this phenomenon. \par



\end{abstract}

\cleardoublepage
\section*{Acknowledgments}
I would like to express my great gratitude to my supervisor Dimitrios Kollias for his supervision and helping me throughout this project. \par

In addition, I would like to thank my family and friends for their love and support.
\clearpage{\pagestyle{empty}\cleardoublepage}

\fancyhead[RE,LO]{\sffamily {Table of Contents}}
\tableofcontents 

\listoffigures
\listoftables

\pagenumbering{arabic}
\setcounter{page}{1}
\fancyhead[LE,RO]{\slshape \rightmark}
\fancyhead[LO,RE]{\slshape \leftmark}

\chapter{Introduction}
Emotions play an important role in every human being's life. Besides language or action \cite{goudelis2013exploring}, all interpersonal communication inherently includes emotions. This interpersonal interaction would be greatly enhanced if we could understand emotions more deeply. \par

Furthermore, since the interactions between human beings and machines are increasingly frequent, machine's understanding of humans emotions benefits the development of Human-Computer Interaction (HCI) technologies. \par

Apart from HCI, automatic Emotion Recognition also has huge potential for applications such as automobile safety, counter-terrorism, psychiatry, medical research \cite{kollias13,tagaris1,tagaris2} and education study \cite{7292443}. \par

Different approaches have been attempted to discover human emotions. \par

George Caridakis et al. \cite{10.1007/978-0-387-74161-1_41} proposed a multimodel approach for recognition of emotions using facial expressions, body movement, gestures and speech. \par

Michel F. Valster and Maja Pantic proposed a fully automatic method which allows not only the recognition of 22 facial muscle actions i.e. action units (AUs), but also temporal characteristics. Temporal characteristics include neutral, onset, offset and etc \cite{6020812}. \par

In this project, Deep Neural Networks are used to achieve emotion recognition. A Convolutional Neural Network (CNN) is a biologically-based trainable architecture that can learn invariant features. It has been proved that CNNs show good performance in image recognition. Recurrent Neural Networks (RNN) are able to reveal the temporal behaviour for time sequence samples, for instance, texts and videos. \par 

By feeding the images containing emotions into the CNN-RNN structure, the image features will be first extracted before RNNs extract temporal features. Fully connected layers are used to make predictions after CNN-RNN architecture.\par

Two different emotion representations are used in this project, i.e. categorical and dimensional. Categorical representation is to classify affect into seven basic emotions, which are happiness, sadness, anger, surprise, fear, disgust and neutral. While, dimensional representation utilizes two continuous values, namely Valence and Arousal, to characterise affect. \par

Three types of CNN-RNN Deep Neural Networks were implemented and fine-tuned to perform the best emotion recognition. The three models are trained on categorical labels, dimensional labels and both respectively. We want to investigate that, by training with two types of labels simultaneously, would the overall performance be increased compared to the single-function models? Also, we want to reveal the inner relationship of two affect representation through this experiment. Extensive experiments have been made to fine-tune the architectures and to find the best performance. \par

Before training the Neural Networks, efforts were made on extending the existing dimensional affect label video database to the one with the categorical emotions. In addition, data preprocessing and organisation were carried out to build relative datasets in order to train and evaluate the models. \par

In conclusion, the aim of the project is to extend a database to one with both emotion category and valence \& arousal annotations. After that, categorical-only, valence-arousal-only and combined models will be trained based on different architectures. The comparison of the performances achieved on different models will be examined, and the inner-relationship of the categorical and dimensional emotion representation will be investigated. \par


\chapter{Background Research}

This chapter introduces the idea of two emotion theories and two Neural Network architectures investigated in the project. Related work in emotion recognition research are also described. \par

Section~\ref{Emotion Theories} discusses the different emotions theory for emotion recognition study. Two representations, emotion categories and valence \& arousal are especially introduced. These two emotion theories are used in this project. Section~\ref{Deep Neural Networks} goes through the idea of the Convolutional Neural Networks and Recurrent Neural Networks. Section~\ref{Related Work and Challenges} demonstrates the related work in emotion recognition study, and some famous emotion recognition competitions. Section 2.4 describes techniques to interpret what neural networks learned. \par

\section{Emotion Theories}\label{Emotion Theories}
In psychology, there are several ways to describe emotions. For instance, there are categorical approaches, dimensional approaches and appraisal-based approaches. In this project, the categorical method and dimensional method will be utilised to examine affect.

\subsection{Basic and Compound Categories}
Ekman and his colleagues conducted a lot of experiments and drew the conclusion that there exist seven basic emotions, namely, happiness, sadness, anger, surprise, fear, disgust and neutral. \cite{FoaEdnaB.1973Eith} \par
The affect can also be classified into multiple basic categories, therefore leading to various compound or complex emotions. Du, Shichuan, and Aleix M. Martinez propose that compound emotions are those that can be constructed by more than one basic component classes to generate new emotions. Their research suggests that emotions are better represented using a series of basic and compound categories rather than only seven basic components \cite{du2015compound}. 15 compound emotions are defined in their study, which are happily surprised, happily disgusted, sadly fearful, sadly angry, sadly surprised, sadly disgusted, fearfully angry, fearfully surprised, fearfully disgusted, angrily surprised, angrily disgusted, disgustedly surprised, appalled, hatred and awed. \par

\begin{figure}[htb]
\centering
\includegraphics[width = 1\hsize]{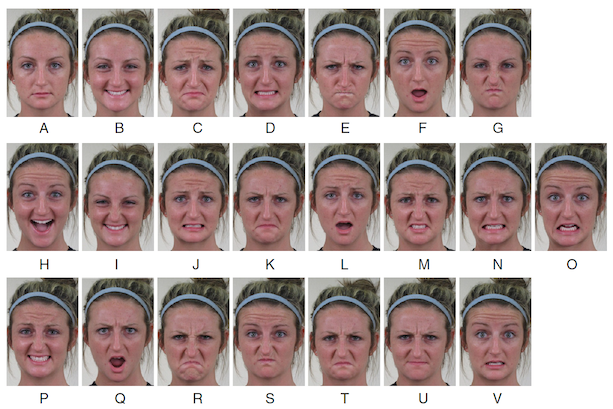}

\caption{Sample images for basic and compound emotions. \cite{du2015compound}}
\label{fig:bc}
\end{figure}

Figure~\ref{fig:bc} shows the seven basic emotions (A to G: neutral, happy, sad, fearful, angry, surprised and disgusted) and 15 compound emotions. For example, emotion H is happily surprised and P is fearfully disgusted \cite{du2015compound}. \par

\subsection{Valence and Arousal Approach}
For dimensional methods, psychologists believe that emotions are correlated to each other in a systematic manner, instead of categorically independent from one another. Through this methodology, the emotion variability can be represented by two dimensions, i.e. Valence (V) and Arousal (A). \par
The valence dimension (V) stands for how positive or negative the affect is. It ranges from unhappy feelings to happy feelings. \par
The arousal dimension (A) stands for how excited the emotion it. Its range is from boredom to excitement. \par
Since the psychological research shows that the valence and arousal are interconnected, duplicate configurations and inter-dependencies within the value exist. \cite{doi:10.1093/cercor/bhk024} \par
Compound emotions can be handled easily in the valence and arousal approach. Because the affect transition can be captured on the dimensions and expression can be accurately and authentically indicated on continuous scales. This is one advantage of the valence and arousal method.

\begin{figure}[htb]
\centering
\includegraphics[width = 0.6\hsize]{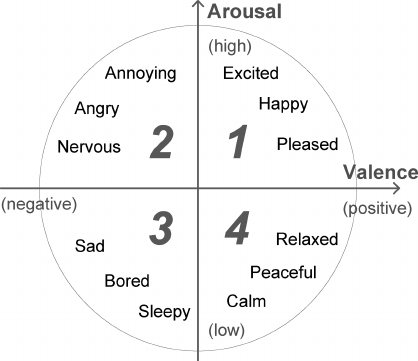}

\caption{A representation of 2D valence-arousal emotion space. \cite{article1}}
\label{fig:logo}
\end{figure}

Figure ~\ref{fig:logo} is a representation of the 2D valence -arousal emotion space. \par

\subsection{Other Emotion Models}
Another famous method to recognise affects is through the activation of muscles on the face. The Facial Action Coding System (FACS) is built based on facial Action Units (AU), which define the activation and contraction of facial muscles. FACS provides the rules for emotion detections of AUs \cite{ekman1997face}.

\subsection{Emotion Representation Relationship Study}
There are researches to discover the relationship of different emotion representations. Stephan Hamann \cite{HAMANN2012458} attempted to map the discrete and dimensional emotions by using neuroimaging technology, to identify brain regions that associated with two types of representations. Sven Buechel and Udo Hahn \cite{buechel2018emotion} presents a neural network to transform valence \& arousal representation to basic emotions for lexicon construction. \par

\section{Deep Neural Networks}\label{Deep Neural Networks}
In this project, a variety of neural networks will be exploited and combined to reach the high performance on the dataset and interpretability will be investigated.

\subsection{Convolutional Neural Networks}
\begin{figure}[htb]
\centering
\includegraphics[width = 1\hsize]{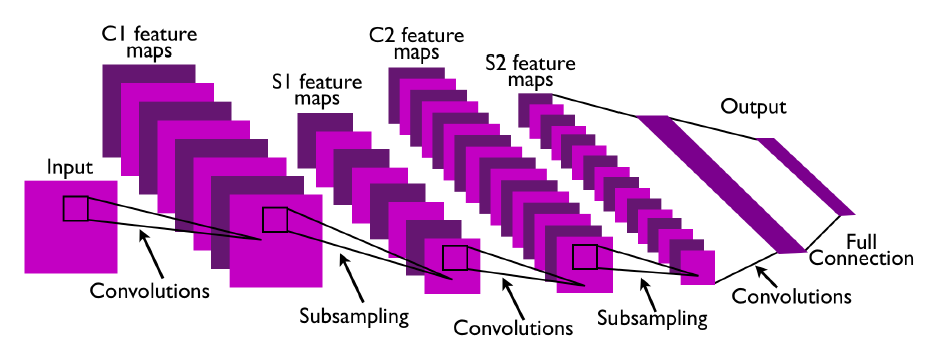}

\caption{A typical architecture of convolutional neural network \cite{5537907}}
\label{fig:convnet}
\end{figure}

Convolutional Neural Networks (CNN) are biologically inspired trainable architectures that can learn invariant features. With the implementation of multiple stages, a CNN is able to learn multi-level hierarchies of features. Each stage in a CNN consists of a convolutional layer and a feature pooling layer, followed by the fully-connected layer to carry out the classification. The input and output of one stage are sets of arrays, namely feature maps \cite{5537907}. Figure~\ref{fig:convnet} is a typical architecture of convolutional neural network. \par

\subsubsection*{Convolutional Layer}
The convolutional layer is composed of a set of filters, which can learn the features. During the forward pass process, the filters are convolved across the input volume and dot products are calculated between the entries of the filter and the input volume at all positions. Each filter recognises a specific feature at input volume spatially. The filters will be learned when they see some type of visual or spatial features. \par
The whole set of filters in one convolutional layer will together produce a feature map as the output, which represents the features extracted from the input. \cite{cs231n} 

\begin{figure}[htb]
\centering
\includegraphics[width = 1\hsize]{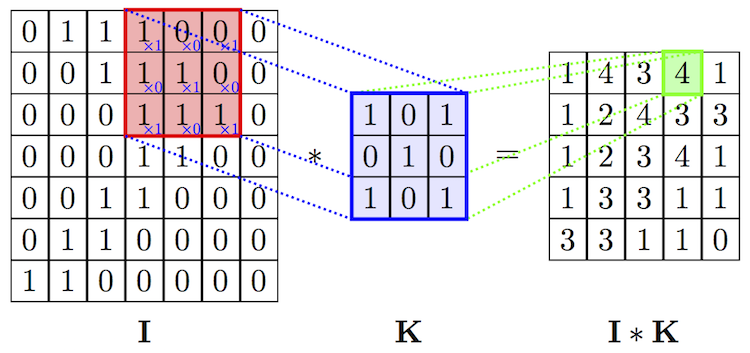}

\caption{A representation of feature extraction in convolutional layer. \cite{spark}}
\label{fig:convo}
\end{figure}

Figure~\ref{fig:convo} shows the convolution process in convolutional layer. \par

\subsubsection*{Pooling Layer}

The function of pooling layer is to decrease the spatial size of the feature maps in order to reduce the number of parameters and computation in the entire neural network. Therefore, the overfitting can be controlled. A pooling layer is usually inserted between a series of convolutional layers. \par

Max operation is the most common sampling operation to spatially resize the feature maps. For example, a pooling layer with filters of size 2x2 and the stride of 2 will pick up the maximum number over 4 numbers. \cite{cs231n}

\begin{figure}[htb]
\centering
\includegraphics[width = 1\hsize]{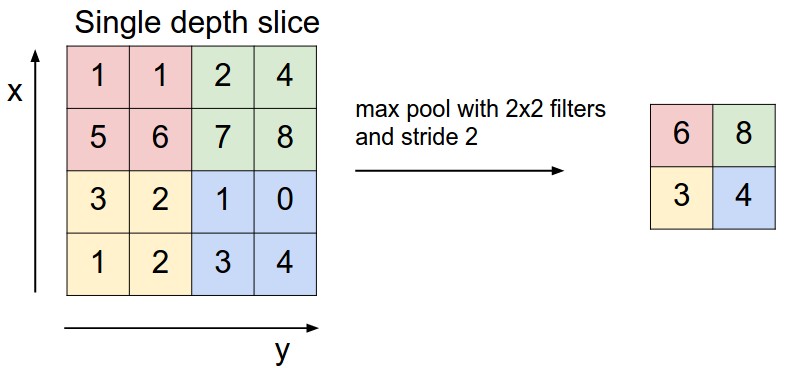}

\caption{A representation of Max Pooling operation. \cite{cs231n}}
\label{fig:maxpool}
\end{figure}

The Max Pooling operation is shown in Figure ~\ref{fig:maxpool}. \par

\subsubsection*{Fully-connected Layer}
The fully-connected layer is a regular neural network. It enables the entire neural network to classify the inputs into categories.

\subsection{Recurrent Neural Networks}
The Recurrent Neural Network (RNN) extends the functions of a regular neural network. It is able to handle a sequence input by having a recurrent hidden stage. The hidden state's activation does not only rely on the input but also the stage of previous time. This can implemented in Equation \ref{equ:recurrent}.

\begin{equation}
{h_t} = g(W{x_t} + U{h_{t-1}})
\label{equ:recurrent}
\end{equation}

where $h_t$ is the recurrent hidden stage, $h_{t-1}$ is the hidden stage of the previous time step, $W$ and $U$ are weight matrix, $g$ is the activation function. \cite{DBLP:journals/corr/ChungGCB14}

\subsubsection*{Long Short-term Memory Unit}
The facts have been observed that it is hard to train recurrent neural networks in order to capture long-term dependencies. The reason is that the gradients are prone to either vanish or explode, which prevents the gradient-minimisation algorithms to find the optimal parameters. \cite{DBLP:journals/corr/ChungGCB14}\par
One popular solution to this problem is to design a more complicated activation function using gate units. Long Short-term Memory unit (LSTM) is one of the attempts. \par

\begin{figure}[htb]
\centering
\includegraphics[width = 1\hsize]{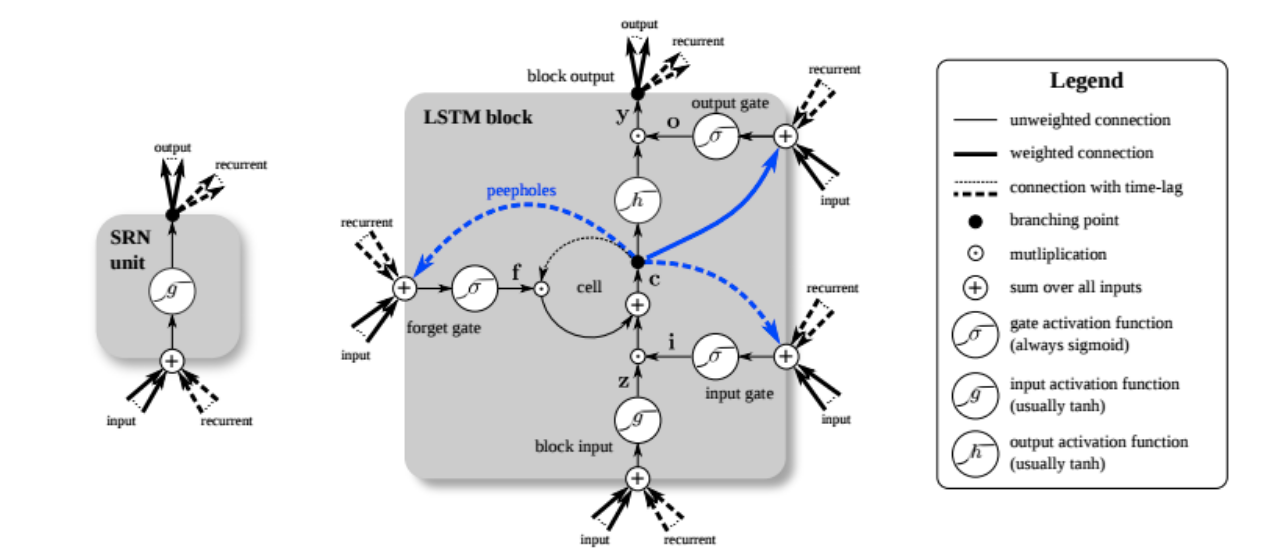}

\caption{A representation of LSTM.\cite{lstm}}
\label{fig:LSTM}
\end{figure}

Different from the regular recurrent neural networks, the LSTM unit does not easily overwrite the recurrent unit state at each time-step. Instead, by introducing three gate units, i.e. input gate, output gate and forgetting gate, LSTM unit can decide whether to reserve the current memory or not. The input gate unit is capable of deciding the degree to which the new memory is added to the memory cell. The output gate unit can determine the degree to which the memory is exposed to outside. And the forget gate unit enables the network to judge the degree to which the current memory is forgotten. The introduction of the gate units can discover the long-term dependencies which are hidden by the short-term dependencies in easy recurrent unit \cite{DBLP:journals/corr/ChungGCB14}. A graphic representation of LSTM can be seen in Figure ~\ref{fig:LSTM}. \par

\section{Related Work and Challenges}\label{Related Work and Challenges}
There are several significant challenges (competitions) that promote the development of emotion recognition. Various approaches were attempted in those challenges. \par

EmotiW focuses on the categorical emotion recognition in-the-wild \cite{Dhall:2017:IGE:3136755.3143004}. In 2017, baseline experiments achieved 38.81\% in the validation set. And the winner won the challenge with 83.9\% validation accuracies using individual facial emotion CNNs and global image based CNNs \cite{Tan:2017:GER:3136755.3143008}. \par

Tan et al. \cite{Tan:2017:GER:3136755.3143008} proposed that individual emotion CNNs extract faces from images before the training. And they used a large-margin softmax loss for discriminative learning with both aligned and non-aligned faces. For the global images based CNNs, they trained the networks with VGC19, BN-Inception and ResNet101 architectures to extract global features for the purpose of recognising group emotion. 

AVEC concentrates on dimensional emotion recognition under controlled environment \cite{nott45489}. The baseline for this challenge is 0.375 and 0.466 Concordance Correlation Coefficient (CCC) for arousal and valence \cite{ringeval2017avec}. The winner won with 0.68 and 0.76 for the two dimensions using a Multi-modal Learning method \cite{Chen:2017:MML:3133944.3133949}. \par

Chen et al. \cite{Chen:2017:MML:3133944.3133949} investigated that the fusion of acoustic, visual and textual modalities and multi-task learning arousal and valence simultaneously can significantly improve the recognition performance. The temporal LSTM-RNN architecture was used in their research.

AffWild \cite{kollias1,kollias2,kollias3,kollias4,kollias5,kollias15} pays attention to dimensional emotion recognition in-the-wild. The baseline architecture using a CNN-M network and obtained CCC of 0.15 and 0.10 for valence \& arousal respectively. And the winning architecture FATAUVA-Net improved the CCC to 0.396 and 0.282. \par

The main idea of FATAUVA-Net architecture is by exploiting Action Units as the mid-level representation for valence and arousal estimation. And the Action Units detection is based on facial attribute recognition. 

One track of the EmotioNet challenge is to research the basic and compound emotion recognition. Last year the winner NTechLab achieved a 94.1\% accuracy compared to baseline 91\% \cite{DBLP:journals/corr/Benitez-QuirozS17}.

\section{Interpretability}

As Machine Learning is being widely applied in the real world, the interpretability and understandability of models are becoming increasingly important \cite{KrakovnaViktoriya2016BIMF,kollias10,simou2008image,kollia2011query,kollias11,simou2007fire,kollias12,glimm2013using,horrocks2011answering}. Debugging  Machine Learning methods could be challenging \cite{43146,kollias6}, as machines could make errors that humans would not \cite{DBLP:journals/corr/PapernotMGJCS16}. This prevents us adopting models in applications which require high accuracy and trust. \par

Maithra Raghu et al. \cite{raghu2017svcca} proposed the technique Singular Vector Canonical Correlation Analysis (SVCCA) to compare representations in different layers and networks. SVCCA discovers the deep representation that layers learned by analysing neuron's activation vector. This allows the comparison of representation between different architectures and the insight into the learning process. \par

Laurens van der Maaten et al. \cite{maaten2008visualizing} present a technique t-Stochastic Neighbor Embedding to compare the similarity of the features that neural network extract. This technique visualise the similarity of datapoints by reducing the high-dimensional representation into 2 dimensions. Hence the interpretation of the data can be observed. \par

Activation Maximisation is proposed by Dumitru Erhan et al. \cite{erhan2009visualizing}. The idea of it is to generate data that can maximumly activate specific neurons. Activation Maximisation allows the research to visualise the interpretation of the features that neural networks learned. \par

Karen Simonyan et al. \cite{simonyan2013deep} addresses the idea to compute the saliency map of the classes for Convolutional Neural Networks. The saliency map allows researchers to observe which part of the images that the classification prediction relies on in 2 dimensional possibility heat maps. \par

\chapter{Methodology}
This chapter describes the approaches to prepare the data for the aim to this project. The design of the architectures and corresponding parameters are also explained in this chapter. \par 

Section 3.1 shows the plan to annotate the database and prepare the training, validation and testing sets. Section 3.2 explains the reason to use Convolutional plus Recurrent Neural Networks and illustrates the three models to build. Section 3.3 introduces the pre-trained CNN networks used in the project. Two objective functions chosen for training the different networks are introduced in section 3.4. Section 3.5 demonstrates both the software and hardware working environment for this project.

\section{Data Preparation}

Because there is no video database with both emotion category labels and valence \& arousal labels, the database used in this project is created by extending an existing emotion recognition database with valence \& arousal annotation to one with category emotion. Time was spent on annotating the videos with emotion categories frame by frame. \par

Data will be processed before fed into neural networks. Videos will need to be captured frame by frame, and faces will be cropped out for training use. Matching the frame and corresponding tables will be developed. Images will be normalized and split into training, validation and testing sets respectively. \par

\section{CNN + RNN Deep Neural Networks}
Convolutional Neural Networks have been widely applied to image recognition and analysis. It has been proved that CNNs can effectively extract spatial features from images. Furthermore, Recurrent Neural Networks exhibit the ability to capture the temporal behaviours from time sequence data. Therefore, to recognise emotion in both categorical and dimensional representation from continuous video frames, CNN plus RNN architectures are used in this project \cite{kollias7,kollias14}.

Image data is fed into CNN to extract spatial information. The extracted features are then fed into RNN for temporal information learning. Fully connected layers are stacked on the top of RNN to give dimensional estimation or classification. \par

\begin{figure}[htb]
\centering
\includegraphics[width = 0.9\hsize]{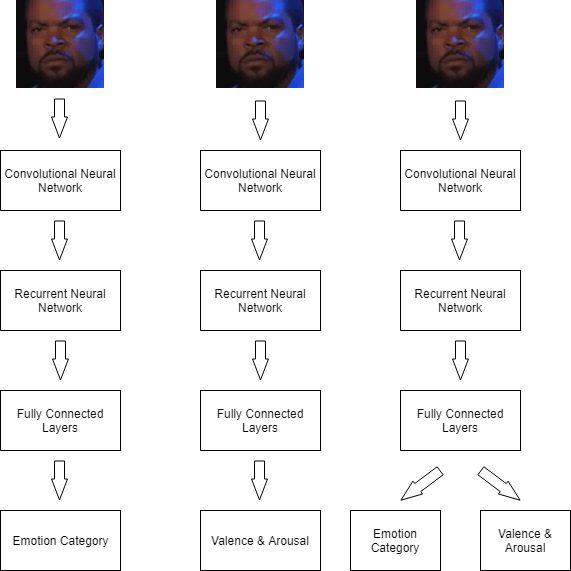}
\caption{A representation of the three models in this project: category-only model, dimensional-only model and combined model}
\label{fig:archi}
\end{figure}

\clearpage

Extensive experiments were carried out based on the three models illustrated in Figure~\ref{fig:archi}. In the first model, we only fed the categorical representation into the CNN-RNN architecture. The second model was only trained on dimensional emotion representation labels. Therefore, the first two models should be able to predict the basic emotion categories or valence \& arousal respectively. \par

Especially, to see if training the model with both categorical and dimensional labels could improve the performance, we built the third combined model. Two emotion representations were fed into the combined model at the same time. The model could learn two representations simultaneously. We want to investigate if this method could lead to a higher performance in two emotion representation recognition. \par

Pre-trained CNN architectures are utilised in this project, including VGG-16, VGG-19 and Xception networks. \par

The pre-trained CNNs are frozen during training. Only the rest of the models are trained and fine-tuned, e.g. hidden layers of RNN and fully connected layers. This operation could increase the flexibility of the architecture, because the further fine-tuning can be implemented by adding new emotion frames or video. \par

\section{Pre-trained Architectures}

Pre-trained models are the architectures that have been trained and hence contains optimal weights and biases for the specific dataset. \par

Training a complicated CNN architecture completely from scratch needs a huge amount of training data, as well as computational power. Using the proved pre-trained models can save training time and features learned from the previous dataset can be transferred to ours. \par

The models have been experimented with various pre-trained CNN architectures. They are all pre-trained on ImageNet database.\par

\subsection{VGG-16 and VGG-19}
The VGG very deep convolutional networks were developed in 2014 at Oxford \cite{DBLP:journals/corr/SimonyanZ14a}. The idea of it is stacking series of  3 x 3 convolutional layers and increasing the depth of Neural Network. It is proved to have great performance on large-scale image recognition.\par

However, one disadvantage of VGG is, it takes a long time to train. This is due to its very deep architecture. \par

The configuration of VGG-16 and VGG-19 used in this project is illustrated in Table ~\ref{table:vgg}. "16" and "19" means the number of weight layers in the architecture. \par

\begin{table}[!htp]
\centering
\begin{tabular}{cc}
\hline
\multicolumn{1}{|c|}{VGG-16}        & \multicolumn{1}{c|}{VGG-19}        \\ \hline
\multicolumn{1}{|c|}{2 * conv3-64}  & \multicolumn{1}{c|}{2 * conv3-64}  \\ \hline
\multicolumn{1}{|c|}{maxpooling}    & \multicolumn{1}{c|}{maxpooling}    \\ \hline
\multicolumn{1}{|c|}{2 * conv3-128} & \multicolumn{1}{c|}{2 * conv3-128} \\ \hline
\multicolumn{1}{|c|}{maxpooling}    & \multicolumn{1}{c|}{maxpooling}    \\ \hline
\multicolumn{1}{|c|}{3 * conv3-256} & \multicolumn{1}{c|}{4 * conv3-256} \\ \hline
\multicolumn{1}{|c|}{maxpooling}    & \multicolumn{1}{c|}{maxpooling}    \\ \hline
\multicolumn{1}{|c|}{3 * conv3-512} & \multicolumn{1}{c|}{4 * conv3-512} \\ \hline
\multicolumn{1}{|c|}{maxpooling}    & \multicolumn{1}{c|}{maxpooling}    \\ \hline
\multicolumn{1}{|c|}{3 * conv3-512} & \multicolumn{1}{c|}{4 * conv3-512} \\ \hline
\multicolumn{1}{|c|}{maxpooling}    & \multicolumn{1}{c|}{maxpooling}    \\ \hline
\multicolumn{1}{|c|}{FC layers}     & \multicolumn{1}{c|}{FC layers}     \\ \hline
\multicolumn{1}{l}{}                & \multicolumn{1}{l}{}              
\end{tabular}
\caption{Configuration of VGG-16 and VGG-19 architecture}
\label{table:vgg}
\end{table}

\subsection{Xception}
\begin{figure}[htb]
\centering
\includegraphics[width = 0.8\hsize]{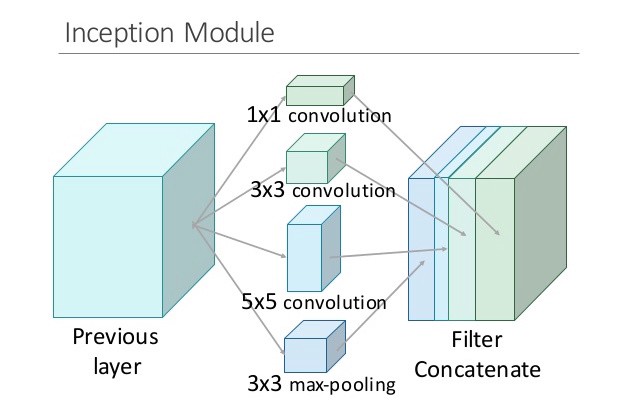}
\caption{A schematic of Inception Module}
\label{fig:inception}
\end{figure}

Xception is an extension of the Inception network. The key idea of the Inception network is combining convolutional layers parallelly. Therefore, each layer is able to concatenate the results from different kernels simultaneously and then feed the result into the next layer as one single output. A demonstration of the Inception Module is shown in Figure~\ref{fig:inception}. \par

\begin{figure}[!htb]
\centering
\includegraphics[width = 0.6\hsize]{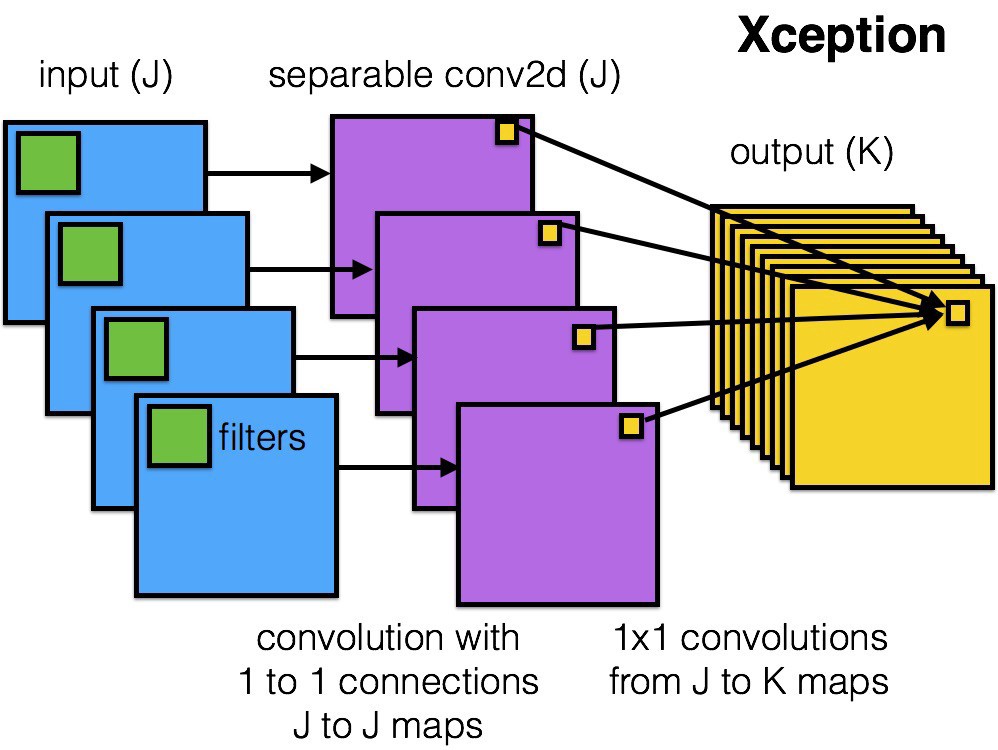}
\caption{Depthwise separable convolutions in Xception}
\label{fig:xception}
\end{figure}

Xception replaces the Inception modules with depthwise separable convolutions, which gives a small improvement in image classification accuracy. The replacement of depthwise separable layers and the demonstration of Xception are illustrated in Figure~\ref{fig:xception}.

\clearpage

\section{Objective Function}
Two types of affect representation are experimented with in this project. The way to evaluate the prediction depends on the representation. Different objective functions are utilised in the neural network training. Categorical cross-entropy and Concordance Correlation Coefficient (CCC) are applied for different output \cite{kollias5}. 

\subsection{Categorical Cross-Entropy}

Cross-entropy function is utilised together with the Softmax classifier as the loss function for the categorical emotion representation. The two functions are defined as follows:

\begin{equation}
Softmax\ Classifier: f_j(z) = \frac{e^{z_j}}{\sum\nolimits_{k}^{} e^{z_k}} 
\end{equation}

\begin{equation}
cross-entropy\ loss: L_i = -log(\frac{e^{f_{yi}}}{\sum\nolimits_{j}^{} e^{f_{j}}})
\end{equation}

where $z_j$ is the output of the fully connected layer for each class.

A Softmax classifier can interpret the scores from the outputs as the unnormalized log probabilities. 

In backpropagation, the equation to calculate the gradient is elegant when using Softmax classifier and Cross-entropy loss. The computational resource is  saved.

\subsection{Concordance Correlation Coefficient (CCC)}

The loss function chosen for Valence \& Arousal was based on the Concordance Correlation Coefficient, which is defined as follows:

\begin{equation}
\rho_c = \frac{2s_{xy}}{s_x^2 + s_y^2 + (\bar{x} - \bar{y}) ^ 2}
\end{equation}

where $\bar{x}$ and $\bar{y}$ are the mean value of the predicted values and the ground truth values. $s_x$ and $s_y$ are the variances respectively. $s_{xy}$ is the respective covariance value.

Concordance Correlation Coefficient measures the agreement between ground truth and prediction, which illustrates a better perception of whether prediction matching the annotation. Consequently, the loss was defined as $1 - \rho_c$.

\section{Software and Hardware Environment}
This project is implemented in the Python language. Libraries including menpo, NumPy, OpenCV and matplotlib are used to process the image data and carry out visualization.

TensorFlow and Keras are utilised to train the model and evaluate the corresponding performance.

Training was accomplished on CSG GPGPU Cluster with Nvidia GeForce GTX Titan X and GTX 1080 Ti.

\chapter{Experiment}
This chapter describes the implemented experiments and work for this project in detail. \par  
Section 4.1 explains how the database is extended. Section 4.2 describes the methods to process the annotated data into training materials. Section 4.3 clarifies how images are resized and normalised before training. Experimented Deep Neural Networks \cite{kollia2009interweaving} are summarised in section 4.4. Section 4.5 illustrates the optimal hyperparameters investigated on different architectures. \par

\section{Annotation}
For data annotation, three weeks were spent on extending the existing valence \& arousal dataset with categorical emotion representation. \par

All videos were first converted to 30 fps (frame per second) for the unity for future cropping process. An Active Unit HTML5 annotation tool was adapted to categorical emotion representation annotation. Original User Interface was modified to make it more intuitive and convenient for categorical emotion usage.\par 

\begin{figure}[!htb]
\centering
\includegraphics[width = 0.8\hsize]{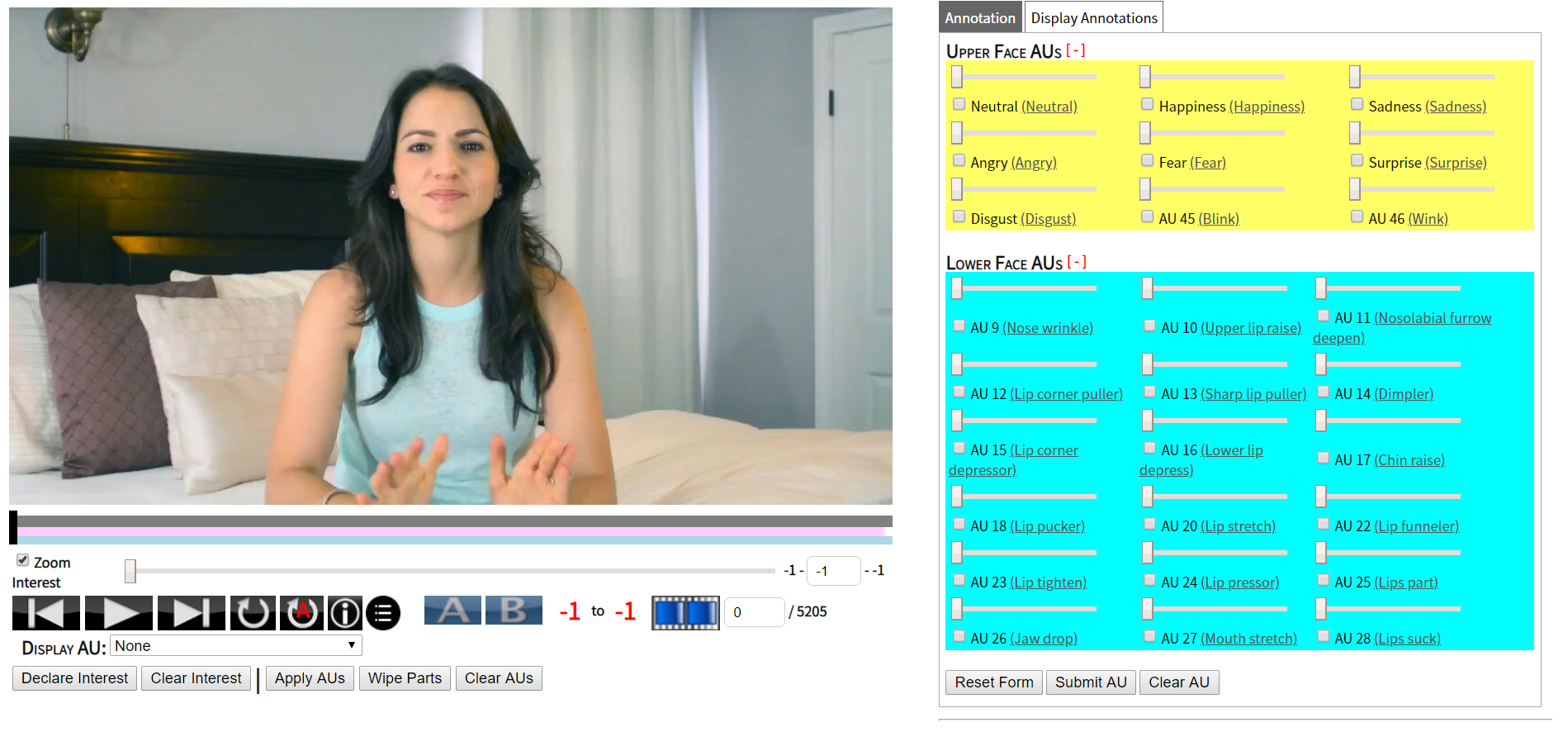}
\caption{Tool used for annotation}
\label{fig:tool}
\end{figure}

Figure~\ref{fig:tool} illustrates the annotation tool.

All the annotation was applied to frames of video in the following process:
\begin{enumerate}
  \item The whole video is watched from beginning to end, hence, the context of video and the atmosphere of the circumstance could be perceived. The possibility of the wrong annotation due to the misleading and complicated facial expression was strongly avoided.
  \item The video is then watched again from beginning to end. The different emotions happened in the video are marked down. 
  \item Each emotion is annotated separately throughout the video to achieve a high-quality representation. 
  \item During annotation, the precise positioning tool is used to annotate the video frame by frame. 
\end{enumerate}

All videos were reviewed at least three times according to the annotation process. Four videos in the dataset have more than one character. Each character in the video had their emotions annotated independently.

\section{Dataset Processing}
The Menpo project \cite{menpo14} was used to extract the faces in all frames of video. ffld2 detector was utilised to detect faces \cite{avrithis2000broadcast} from the frame~\cite{10.1007/978-3-319-10593-2_47}. For videos with more than one character, a software was written to classify different faces based on the coordinates of the faces in the frame. Images were then reviewed and corrected manually. 265,663 faces were cropped from the videos. \par

After cropping the faces out of the frames, matching processes were developed to build the new dataset. The provided valence \& arousal annotations has several formats. A software was developed to translate them into one unifying schematic. This makes it possible to feed the data and labels to the Neural Network. \par  

The cropped faces were then matched with the annotation timestamps by a binary search based nearest neighbour method. For each frame time face instance, the nearest neighbour is searched and corresponding valence \& arousal values are allocated to the face. 

After further matching the frame with the emotion categories, faces without a categorical emotion representation label were removed from the dataset for category-only models and combined models. 107,640 frames have an emotion category in the database. \par

Because all frames have a pair of valence and arousal values, if we train the valence \& arousal only models on all the frames, and train category-only and combined models on images only having the emotion categories, the size of the database could influence performance. Especially, when we compare the valence \& arousal performance between dimensional-only models and combined models, it can be the size of the database that impacts the performance, rather than the architecture. \par

Therefore, all the three models were trained on the images in the dataset that have both categorical emotion and valence \& arousal labels. The models to predict categorical emotions, valence \& arousal and both of them, would be trained on the single dataset. The variable of the database size is eliminated. \par

\begin{figure}[htp]
\centering
\begin{minipage}[t]{0.25\textwidth}
\centering
\captionsetup{width=0.95\textwidth}
\includegraphics[width=\textwidth]{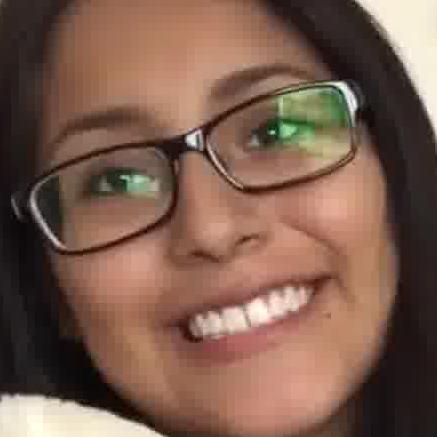}
\caption{\small A representation of the frame in Dataset: Happy, Valence: 0.53, Arousal: 0.62}
\label{fig:sam_happy}
\end{minipage}
\begin{minipage}[t]{0.25\textwidth}
\centering
\captionsetup{width=0.95\textwidth}
\includegraphics[width=\textwidth]{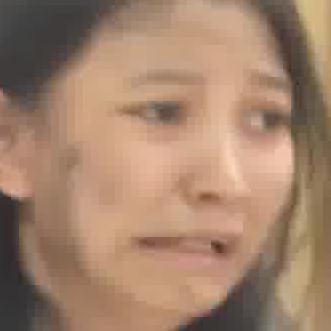}
\caption{\small A representation of the frame in dataset: Fear, Valence: -0.68, Arousal: 0.72}
\label{fig:sam_fear}
\end{minipage}
\end{figure}

In total, now the dataset has 55 videos and is with 107,640 frames. All frames have both dimensional emotion prediction of arousal and valence, as well as categorical annotation such as happy and sad. Figure ~\ref{fig:sam_happy} and Figure ~\ref{fig:sam_fear} illustrate two typical samples and corresponding labels in the dataset.\par

\begin{figure}[htp]
\centering

\begin{minipage}[t]{.48\textwidth}
\centering
\captionsetup{width=1\textwidth}
\includegraphics[width=\textwidth]{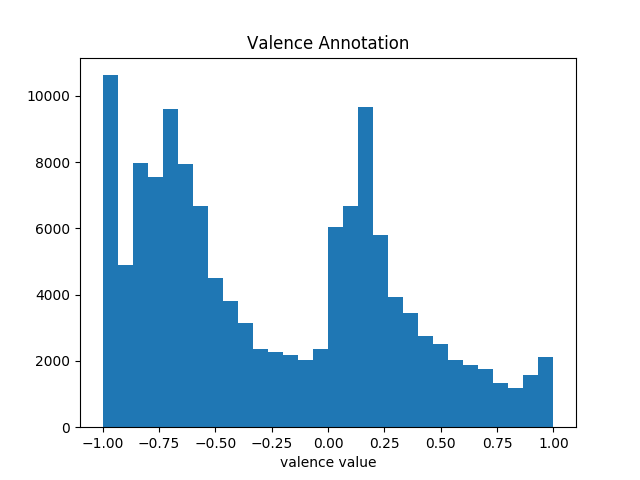}
\caption{\small Valence Value Distribution in Database}
\label{fig:data_all}
\end{minipage}
\begin{minipage}[t]{.48\textwidth}
\centering
\captionsetup{width=1\textwidth}
\includegraphics[width=\textwidth]{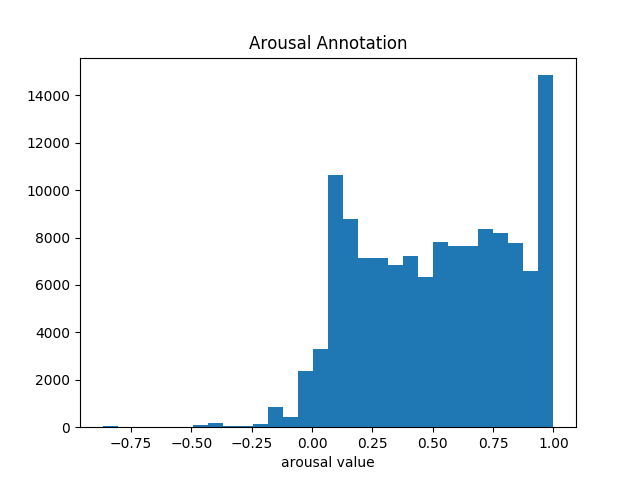}
\caption{\small Arousal Value Distribution in Database}
\label{fig:data2_all}
\end{minipage}

\begin{minipage}[t]{1\textwidth}
\centering
\captionsetup{width=1\textwidth}
\includegraphics[width=\textwidth]{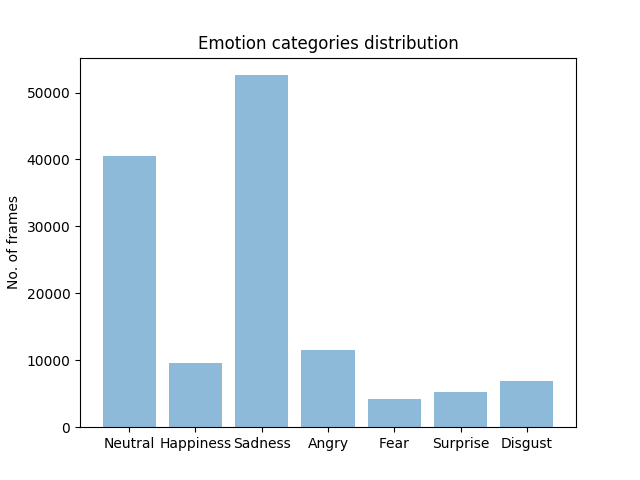}
\caption{\small Categorical Emotion Distribution in Database}
\label{fig:data3_all}
\end{minipage}\hfill
\end{figure}

\begin{figure}[htp]
\centering

\begin{minipage}[t]{.48\textwidth}
\centering
\captionsetup{width=1\textwidth}
\includegraphics[width=\textwidth]{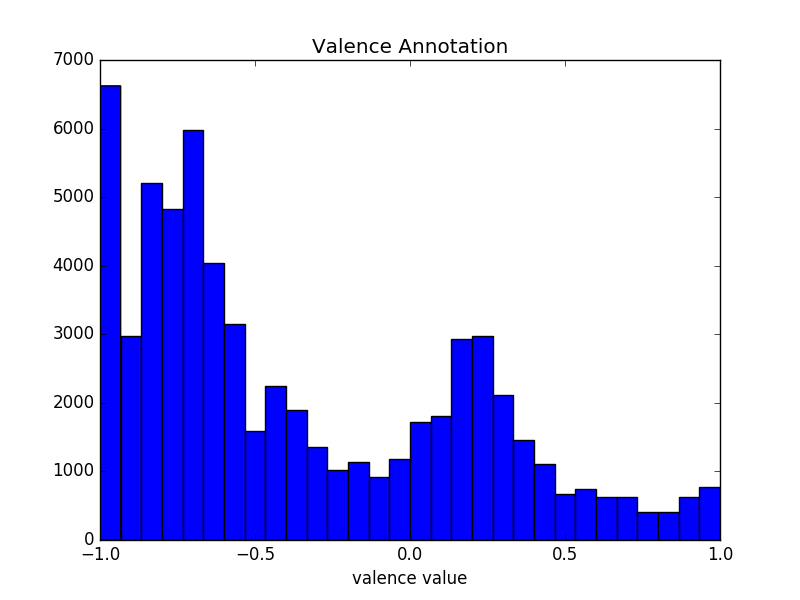}
\caption{\small Valence Value Distribution in Training Set}
\label{fig:data1}
\end{minipage}
\begin{minipage}[t]{.48\textwidth}
\centering
\captionsetup{width=1\textwidth}
\includegraphics[width=\textwidth]{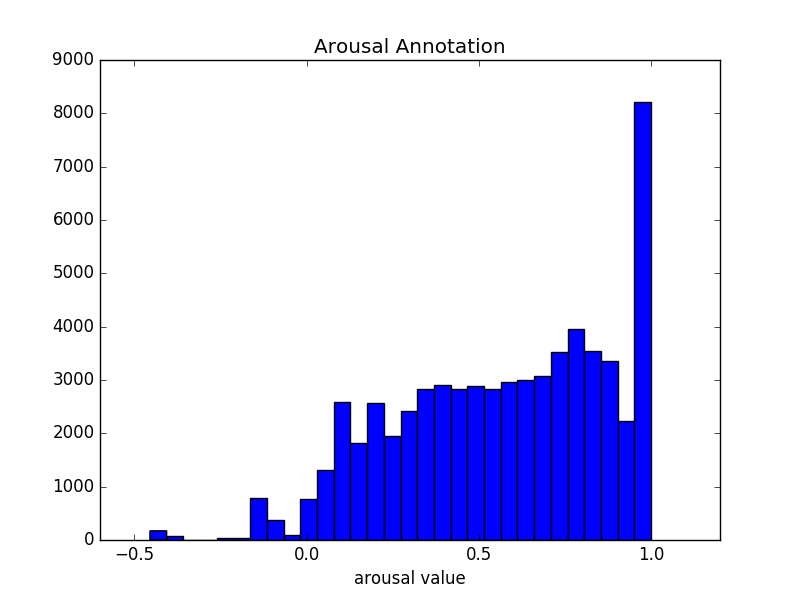}
\caption{\small Arousal Value Distribution in Training set}
\label{fig:data2}
\end{minipage}

\begin{minipage}[t]{1\textwidth}
\centering
\captionsetup{width=1\textwidth}
\includegraphics[width=\textwidth]{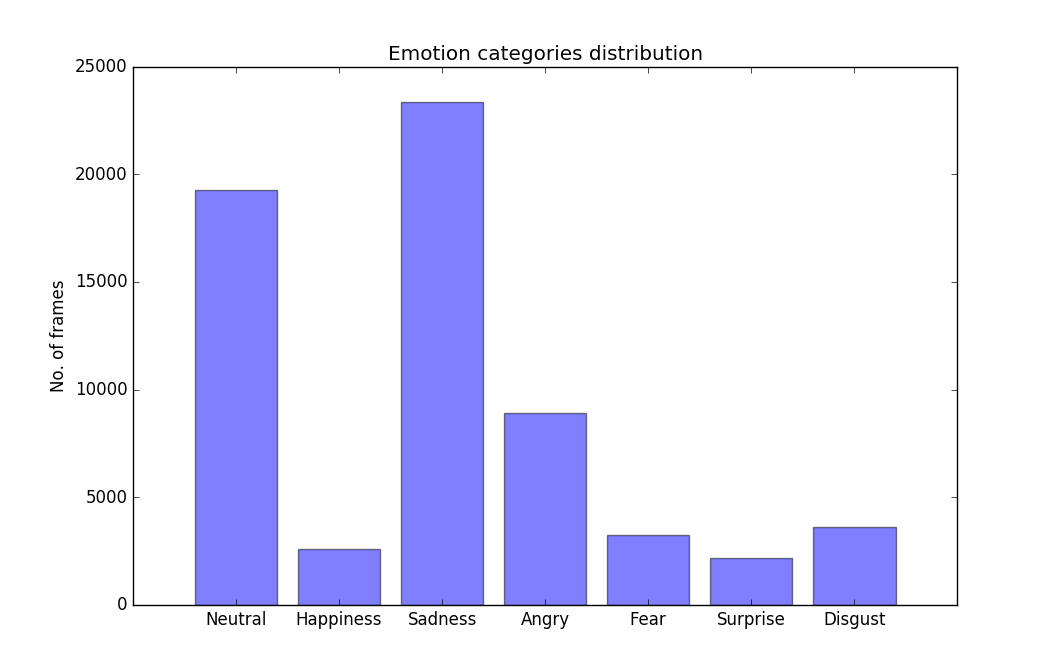}
\caption{\small Categorical Emotion Distribution in Training Set}
\label{fig:data3}
\end{minipage}\hfill
\end{figure}

\begin{figure}[htp]
\centering
\begin{minipage}[t]{.48\textwidth}
\centering
\captionsetup{width=1\textwidth}
\includegraphics[width=\textwidth]{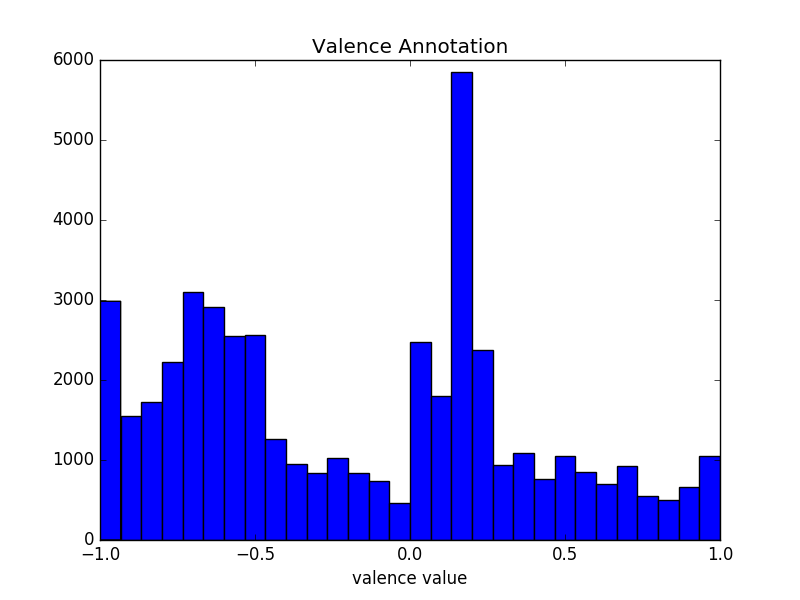}
\caption{\small Valence Value Distribution in Validation Set}
\label{fig:data4}

\end{minipage}
\begin{minipage}[t]{.48\textwidth}
\centering
\captionsetup{width=1\textwidth}
\includegraphics[width=\textwidth]{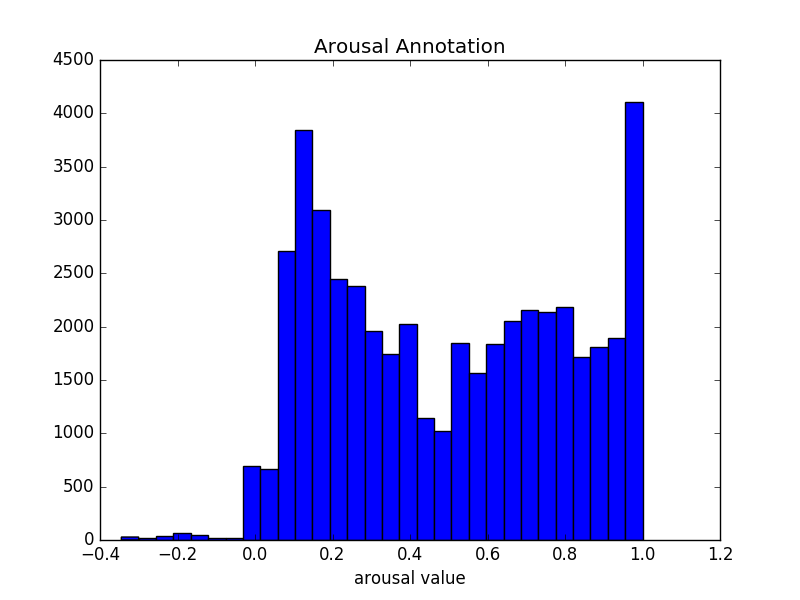}
\caption{\small Arousal Value Distribution in Validation Set}
\label{fig:data5}
\end{minipage}

\begin{minipage}[t]{1\textwidth}
\centering
\captionsetup{width=1\textwidth}
\includegraphics[width=\textwidth]{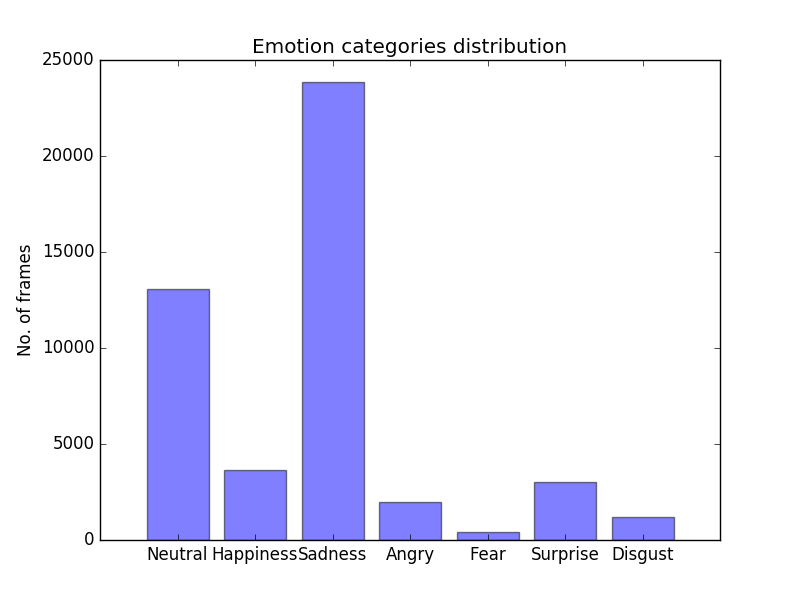}
\caption{\small Categorical Emotion Distribution in Validation Set}
\label{fig:data6}
\end{minipage}\hfill
\end{figure}

\begin{figure}[!htp]
\centering
\begin{minipage}[t]{.48\textwidth}
\centering
\captionsetup{width=1\textwidth}
\includegraphics[width=\textwidth]{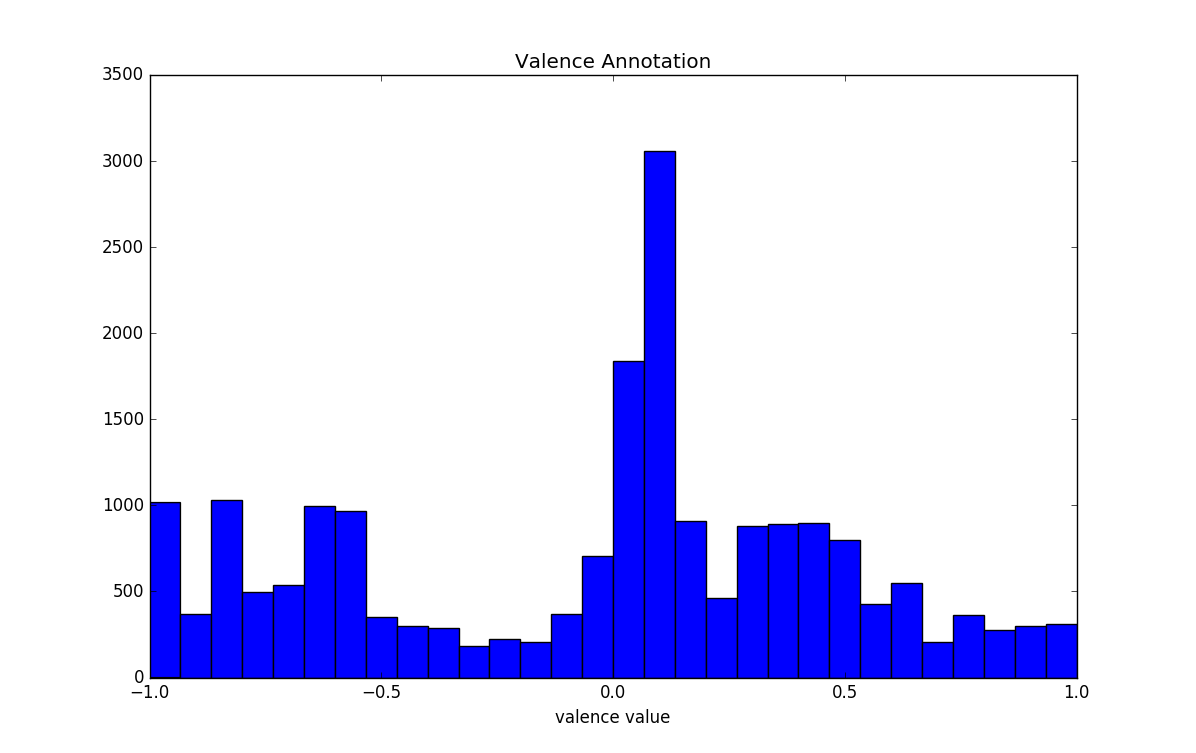}
\caption{\small Valence Value Distribution in Testing Set}
\label{fig:data7}

\end{minipage}
\begin{minipage}[t]{.48\textwidth}
\centering
\captionsetup{width=1\textwidth}
\includegraphics[width=\textwidth]{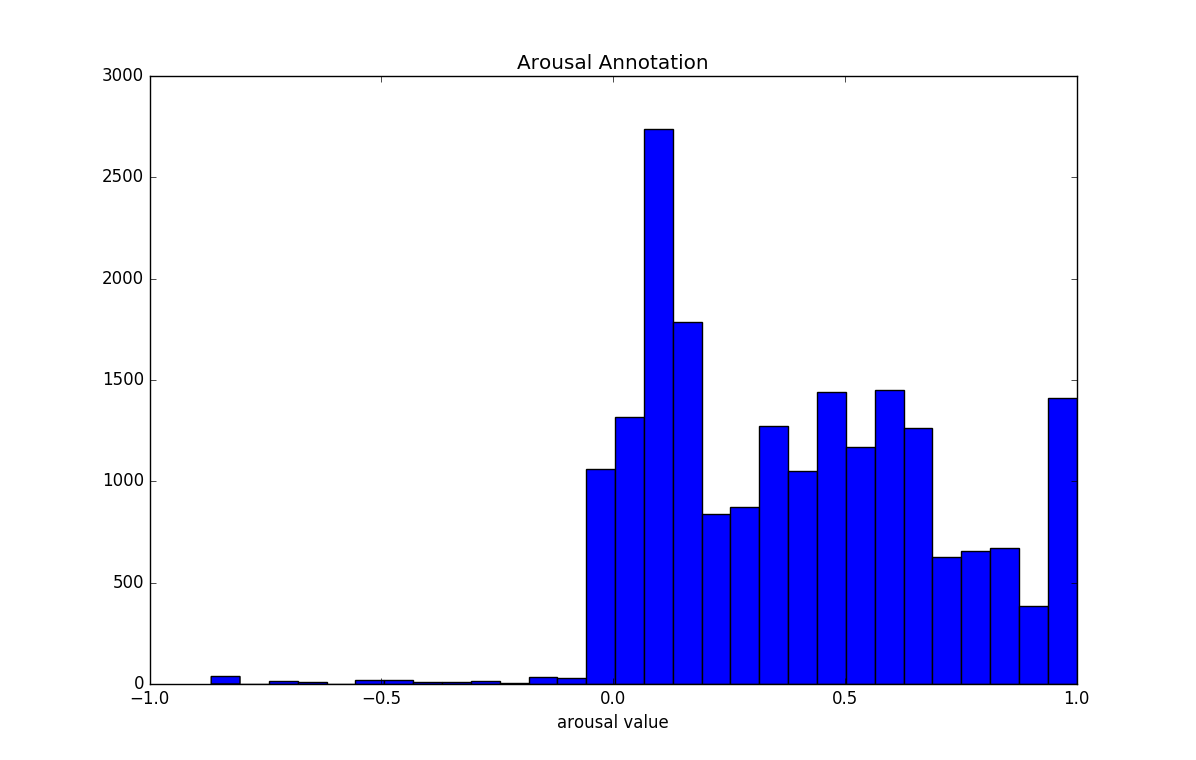}
\caption{\small Arousal Value Distribution in Testing Set}
\label{fig:data8}
\end{minipage}

\begin{minipage}[t]{1\textwidth}
\centering
\captionsetup{width=1\textwidth}
\includegraphics[width=\textwidth]{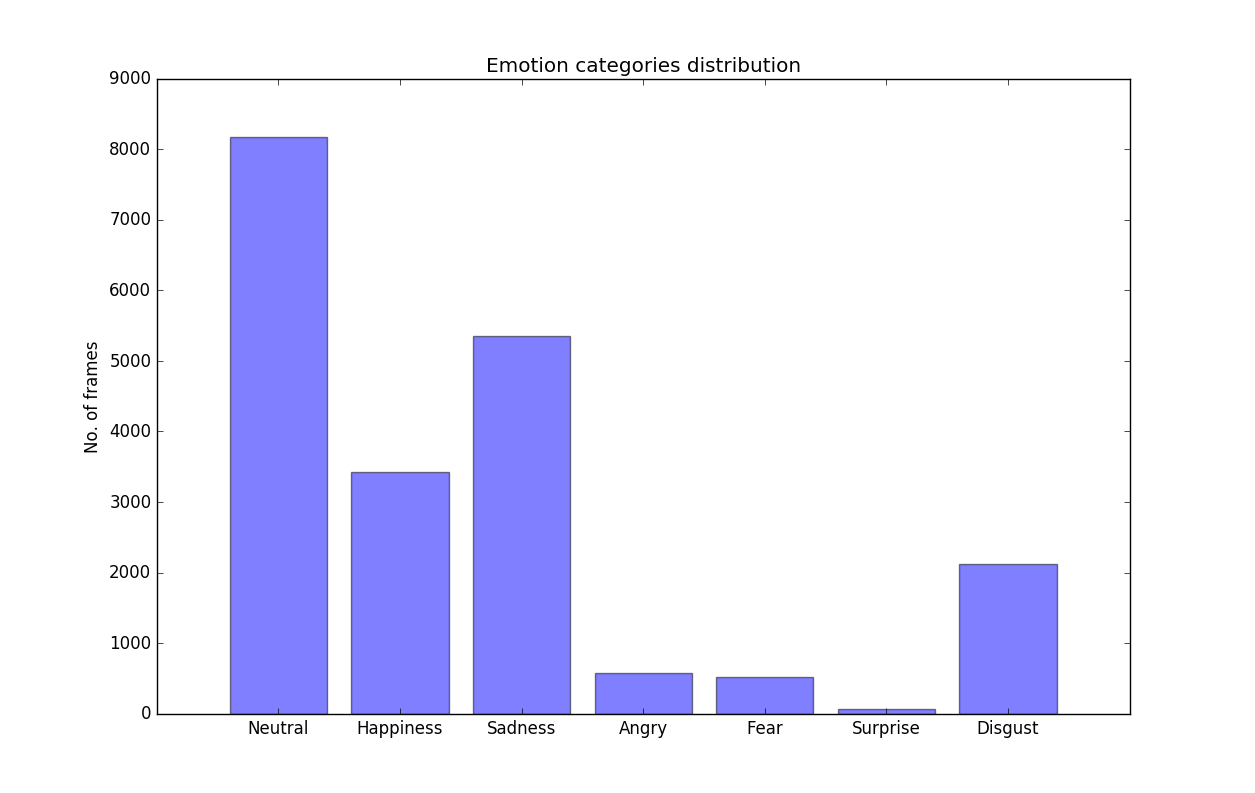}
\caption{\small Categorical Emotion Distribution in Testing Set}
\label{fig:data9}
\end{minipage}\hfill
\end{figure}

All videos with corresponding frames were split into three datasets, i.e. training, validation and testing. The videos were randomly selected and put into each dataset first, followed by manual adjustment to try and ensure that all three datasets have every label kind, although this could not be guaranteed. Figure~\ref{fig:data_all} - Figure~\ref{fig:data9} provide histograms for the annotated values of categorical emotions and valence and arousal in the database and various sets. It can be observed that there are limited negative arousal values in the dataset. \par

\clearpage

It is also worth noting from them that:

\begin{enumerate}
  \item The database for emotion categories are highly imbalanced. Neutral and sadness account for more than 80\%. The effect of the imbalance and solution to address it will be discussed in analysis and future work. 
  \item The level of representation of the validation and testing set for valence and arousal values is not as good as the emotion categories. Because there are only 55 videos in the database, and we split the videos both video-based and person-independent to achieve an impartial evaluation, it is extremely hard to ensure that all the sets' label distributions are similar. What's more, when splitting the videos, there are two types of labels need to take into account. Emotion category and valence \& arousal values are both correlated and independent to some extent. When one labels completeness and level of representation is fulfilled,  the other one would not be as good. The effect of this issue and solutions will also be addressed in analysis and future work. 
  \item The sets have been re-split and evaluated several times, however, this is the best distribution achieved due to the constraints. 
\end{enumerate}

\section{Data Pre-processing}

Before being fed to the end-to-end Deep Neural Network, all frame images were resized to 72 x 72 x 3 pixel resolution. Images intensity values were normalised to a range of [-1, 1]. \par

\clearpage

\clearpage

\clearpage

\clearpage

\section{Deep Neural Networks}

There are different pre-trained CNN architectures utilised in this project. In addition, different fully connected layers are tested for three types of models. \par

\subsection{Category-Only Models}

\begin{table}[htp]
\centering
\begin{tabular}{|l|l|l|}
\hline
\multicolumn{1}{|c|}{block1} & \multicolumn{1}{c|}{VGG-16 or Xception or VGG-19 conv\&pooling parts}                       & \multicolumn{1}{c|}{}          \\ \hline
\multicolumn{1}{|c|}{block2} & \multicolumn{1}{c|}{\begin{tabular}[c]{@{}c@{}}LSTM RNN layer\\ dropout layer\end{tabular}} & \multicolumn{1}{c|}{Unit: 128} \\ \hline
\multicolumn{1}{|c|}{block3} & \multicolumn{1}{c|}{fully connected layer + SoftMax activation}                                                 & \multicolumn{1}{c|}{Unit: 7}   \\ \hline                                                          
\end{tabular}
\caption{Different 1 fully connected layer architectures attempted for category-only models}
\label{table:cate1}
\end{table}

\begin{table}[htp]
\centering
\begin{tabular}{|c|c|c|}
\hline
block1 & VGG-16 or Xception or VGG-19 conv\&pooling parts                              &                                                                                           \\ \hline
block2 & \begin{tabular}[c]{@{}c@{}}LSTM RNN layer\\ dropout layer\end{tabular}        & Unit: 128                                                                                 \\ \hline
block3 & \begin{tabular}[c]{@{}c@{}}fully connected layer\\ dropout layer\end{tabular} & \begin{tabular}[c]{@{}c@{}}Unit: 64 or 128\end{tabular} \\ \hline
block4 & \begin{tabular}[c]{@{}c@{}}fully connected layer + SoftMax activation\end{tabular}       & Unit: 7                                                                                   \\ \hline
\end{tabular}
\caption{Different 2 fully connected layers architectures attempted for category-only models}
\label{table:cate2}
\end{table}

For the category-only models, VGG-16, Xception and VGG-19 have been selected as CNN part. LSTM were used as RNN networks. The hidden unit chosen for LSTM is 128. The activation function of the last fully connected layers is SoftMax, used with the categorical-crossentropy loss function. A representation of these architectures is summarised in Table~\ref{table:cate1}. \par

Furthermore, the same architectures with a different number of fully connected layers and hidden units are experimented. The architectures achieved the best performance for VGG-16 is one more fully connected layers with 64 hidden units. And for VGG-16 and Xception, the additional fully connected layer is the one with 128 units. The configurations of those 2 fully connected layers architectures are summarised in Table~\ref{table:cate2}. \par

\subsection{Dimensional-Only Models}

For the valence \& arousal architectures, the chosen architectures are similar to the category-only models. VGG-16, Xception and VGG-19 plus LSTM architecture are implemented. The difference is the hidden unit for the last fully connected layer is 2 for the purpose of valance and arousal prediction. And the corresponding activation function for the last layer is a linear function. \par

\subsection{Combined Models}

For the implementation of combined models, a single-input multi-output model is utilised. The sequences of images are fed as input, and the two outputs are emotion category and valence \& arousal values respectively. The predictions are based on the shared CNN + RNN layers. VGG-16, Xception and VGG-19 are concatenated with LSTM as the shared CNN + RNN architecture. Different fully connected layers are used for affect category and dimensional representation separately. \par

For VGG-16 based combined models, 7 units fully connect layer for categorical prediction and 2 units for dimensional are applied. Furthermore, 3 fully connected layers for each output are tested e.g. 128 unit fully connected layer + 128 unit fully connected layer + 7 unit fully connected layer for category and 128 unit fully connected layer + 128 unit fully connected layer + 2 unit fully connected layer for dimensional output. \par

For Xception based combined architectures, 128 unit fully connected layer + 7 unit fully connected layer and 128 unit fully connected layer + 2 unit fully connected layer are used for categories and dimensions respectively at first. Three fully connected layers are also tried for this pre-trained network. The three fully connected layers are the same as the ones shown in the last paragraph. \par

In the VGG-19 based combined models, 7 unit fully connected layer and 128 unit fully connected layer + 7 unit fully connected layer are attempted for emotion categories. 2 unit fully connected layer and 128 unit fully connected layer + 2 unit fully connected layer are implemented for valence and arousal prediction. Table~\ref{table:combined} is a summary of the combined models' architectures. \par

\begin{table}[htp]
\centering
\begin{tabular}{|c|c|c|}
\hline
block1     & \begin{tabular}[c]{@{}c@{}}shared CNN + RNN layers\\ dropout layer\end{tabular} &                                                                \\ \hline
block2 - 1 & \begin{tabular}[c]{@{}c@{}}fully connected layers\\ dropout layer\end{tabular}        & \begin{tabular}[c]{@{}c@{}}Unit: \\ 7/2\end{tabular}             \\ \hline
block2 - 2 & \begin{tabular}[c]{@{}c@{}}fully connected layers\\ dropout layer\end{tabular}     & \begin{tabular}[c]{@{}c@{}}Unit: \\ 128 + 7/2\end{tabular}       \\ \hline
block2 - 3 & \begin{tabular}[c]{@{}c@{}}fully connected layers\\ dropout layer\end{tabular}     & \begin{tabular}[c]{@{}c@{}}Unit: \\ 128 + 128 + 7/2\end{tabular} \\ \hline
\end{tabular}
\caption{Summary of the combined model architectures}
\label{table:combined}
\end{table}

\section{Hyperparameters}
Apart from the pre-trained neural networks and fully connected layer, extensive experiments have been developed by choosing different hyperparameters, including: \par

\begin{enumerate}
  \item batch size for the neural network to update
  \item the value of the learning rate
  \item the value of the learning rate decay
  \item the value of the dropout possibility
\end{enumerate}

The sequence length of frames is tested with the value 60 and 80. The best result has been obtained when sequence length is 60. \par

Batch size was chosen in the range 1 to 4. The best performance achieved with batch size 2. \par

Dropout possibility was tested in the range 0.5 to 0.8, and the optimal found was 0.5. \par

Learning rate was selected from 0.1 to 0.000001. The best performance achieved is based on the various architectures. In category-only models and combined models, best results have been achieved with learning rate 0.00001. In the valence and arousal models, 0.0001 have been found to reach the best evaluation values. \par

Similarly, the best learning rate decay varies on different architectures. In category-only architectures and valance \& arousal architectures, the best results were observed when the value of learning rate decay is 0.0001. In the combined models, the value of decay performs the best result is 0.00001. \par

Another parameter to mention is the loss weight in combined models. Since the combined models are multi-output architectures, two loss functions mentioned in Methodology are both applied in the neural networks. Weights are needed to assign to each function. The loss of categorical output is much larger than the loss of dimensional output due to the different definitions. 0.33 for categorical cross-entropy and 1.00 for CCC was found to achieve the best overall performance. \par

In conclusion, extensive experiments were implemented to achieve the best performances. All the results performed by the architectures mentioned with the optimal parameters will be displayed and discussed in the next chapter. \par

\chapter{Result \& Analysis}
This chapter demonstrates all the results obtained from the designed architectures described in the last chapter. Comparisons and analysis are given in this chapter. \par

Section 5.1 and section 5.2 present the performance obtained by category-only models and valence \& arousal-only models on the validation set. Section 5.3 describes the evaluation of the combined models on the validation set. Comparisons to single-functional models are stated in this section. Section 5.4 discovers the inner-relationship of two types of affect representations. Analysis of the impact of the inner-relationship is presented in section 5.5. Section 5.6 discusses the two techniques utilised to discover the interpretability of the neural networks. \par

The evaluation presented in this chapter is performed on validation set. The evaluation on training set and testing set are attached in Appendix. "FC" in the tables is the abbreviation of fully connected layer. \par

\section{Categorical-Only Models}

\begin{table}[htb]
\centering
\begin{tabular}{|c|c|c|c|c|c|c|c|c|}
\hline
                                           & \multicolumn{8}{c|}{Accuracy}                                              \\ \hline
                                           & Neutral & Happy & Sad  & Angry & Fear & Surprise & Disgust & Total         \\ \hline
\cellcolor[HTML]{EFEFEF}VGG16+LSTM+1FC          & 0.08    & 0.00  & 0.77 & 0.00  & 0.00 & 0.00     & 0.00    & 0.35          \\ \hline
\cellcolor[HTML]{EFEFEF}\textbf{VGG16+LSTM+2FC}          & 0.29    & 0.00  & 0.86 & 0.02  & 0.05 & 0.00     & 0.09    & \textbf{0.47}          \\ \hline
\cellcolor[HTML]{EFEFEF}Xcep+LSTM+1FC           & 0.24    & 0.00  & 0.50 & 0.00  & 0.00 & 0.00     & 0.00    & 0.32          \\ \hline
\cellcolor[HTML]{EFEFEF}Xcep+LSTM+2FC           & 0.42    & 0.00  & 0.44 & 0.00  & 0.00 & 0.00     & 0.00    & 0.34          \\ \hline
\cellcolor[HTML]{EFEFEF}VGG19+LSTM+1FC & 0.02    & 0.00  & 1.00 & 0.00  & 0.00 & 0.00     & 0.00    & 0.51 \\ \hline
\cellcolor[HTML]{EFEFEF}VGG19+LSTM+2FC          & 0.26    & 0.00  & 0.24 & 0.00  & 0.00 & 0.00     & 0.00    & 0.19          \\ \hline
\end{tabular}
\caption{Accuracies obtained on validation set by various CNN-RNN architectures for categorical-only models}
\label{table:cate_res}
\end{table}

The accuracies of seven basic emotions obtained on validation set by different attempted architectures are summarised in Table~\ref{table:cate_res}. 

The best accuracy performance was achieved by VGG19 + LSTM + 1 fully connected layers. However, the VGG16 + LSTM + 2 fully connected layers were selected as the best architecture. This is because this model achieved second highest overall accuracy, which is 47\%. Although the overall accuracy 47\% is slightly lower than VGG19's 51\%, it can predict more emotions than VGG19 model. The average accuracy obtained by six architectures was 0.37. \par

It can be observed that, emotions like happiness, angry, fear, surprise and disgust could not be well recognised by the model. This is most likely because of the imbalanced dataset trained on. Emotion neutral and sadness account for more than 80\% of the whole dataset. \par

The combined model is expected to alleviate this problem. Several solutions to solve this problem will be discussed in future work. \par

\section{Valence \& Arousal-Only Models}

Table~\ref{table:va} illustrates the Concordance Correlation Coefficient and mean square errors of predicted Valence \& Arousal values. The best performance is achieved by VGG-19 + LSTM + 2 fully connected layers, which has 128 units and 2 units respectively. \par

The best CCC accomplished is 0.16 for valence and 0.17 for arousal in this VGG-19 architecture. \par

\begin{table}[htb]
\centering
\begin{tabular}{|c|c|c|c|c|}
\hline
                                    & \multicolumn{2}{c|}{CCC}      & \multicolumn{2}{c|}{MSE}      \\ \hline
                                    & Valence       & Arousal       & Valence       & Arousal       \\ \hline
\cellcolor[HTML]{EFEFEF}VGG16+LSTM+1FC & 0.04          & 0.02         & 0.40          & 0.14          \\ \hline
\cellcolor[HTML]{EFEFEF}VGG16+LSTM+2FC & 0.09          & 0.09          & 0.32          & 0.11          \\ \hline
\cellcolor[HTML]{EFEFEF}Xcep+LSTM+1FC  & -0.03          & -0.13          & 0.31          & 0.12          \\ \hline
\cellcolor[HTML]{EFEFEF}Xcep+LSTM+2FC  & -0.01          & -0.03          & 0.30          & 0.10          \\ \hline
\cellcolor[HTML]{EFEFEF}VGG19+LSTM+1FC & 0.11          & 0.14          & 0.32          & 0.11          \\ \hline
\cellcolor[HTML]{EFEFEF}\textbf{VGG19+LSTM +2FC} & \textbf{0.16} & \textbf{0.17} & \textbf{0.31} & \textbf{0.09} \\ \hline
\end{tabular}
\caption{CCC and MSE evaluation of valence \& arousal predictions reached by the CNN-RNN architectures in dimensional-only models on validation set}
\label{table:va}
\end{table}

\section{Combined Models}

Because the combined models are multi-output neural networks. The input images share the CNN and RNN part and have different fully connected layers to make the respective decision. The performance for different output are given separately. Comparisons are made to the categorical-only model and valence \& arousal-only model individually. \par

\section*{Categorical Predictions Evaluation}

\begin{table}[htb]
\centering
\begin{tabular}{|c|c|c|c|c|c|c|c|c|}
\hline
                                           & \multicolumn{8}{c|}{Accuracy}                                              \\ \hline
                                           & Neutral & Happy & Sad  & Angry & Fear & Surprise & Disgust & Total         \\ \hline
\cellcolor[HTML]{EFEFEF}VGG16+LSTM+1FC          & 0.30    & 0.00  & 0.90 & 0.01  & 0.01 & 0.00     & 0.00    & 0.54          \\ \hline
\cellcolor[HTML]{EFEFEF}VGG16+LSTM+3FC          & 0.32    & 0.00  & 0.97 & 0.00  & 0.00 & 0.00     & 0.00    & 0.57          \\ \hline
\cellcolor[HTML]{EFEFEF}Xcep+LSTM+2FC           & 0.09    & 0.00  & 0.81 & 0.00  & 0.00 & 0.00     & 0.00    & 0.43          \\ \hline
\cellcolor[HTML]{EFEFEF}Xcep+LSTM+3FC           & 0.23    & 0.00  & 0.89 & 0.00  & 0.00 & 0.00     & 0.00    & 0.51          \\ \hline
\cellcolor[HTML]{EFEFEF}\textbf{VGG19+LSTM+1FC} & 0.39    & 0.00  & 0.93 & 0.15  & 0.25 & 0.00     & 0.00    & \textbf{0.59} \\ \hline
\cellcolor[HTML]{EFEFEF}VGG19+LSTM+2FC          & 0.36    & 0.00  & 0.84 & 0.11  & 0.18 & 0.00     & 0.00    & 0.53          \\ \hline
\end{tabular}
\caption{Accuracies obtained on validation set by various CNN-RNN architectures for combined models}
\label{table:combined_cate}
\end{table}

The category accuracy results which combined models obtained are summarised in Table~\ref{table:combined_cate}. A significant increase in overall accuracy can be observed. The average accuracy for all the combined models is 0.53, which is higher than the best result shown in the category only model (0.51). \par

The best accuracy achieved is 0.59, increased by 25\% compared to the category-only model. This accuracy is performed on the combined model which has the shared VGG-19 + LSTM, and 1 fully connected layer (7 units). Table ~\ref{table:compare_cate} shows the comparison of the best accuracies obtained on the validation set by combined model and category-only model.\par

It is worth noticing that, even trained and validated on the same very imbalanced dataset, the neural networks shows the ability to learn the minority emotions in some combined models. Emotion angry and fear's accuracy is improved to  0.15 and 0.25 from around 0.00. \par

\begin{table}[htb]
\centering
\begin{tabular}{|c|c|c|c|c|c|c|c|c|}
\hline
                                           & \multicolumn{8}{c|}{Accuracy}                                              \\ \hline
                                           & Neutral & Happy & Sad  & Angry & Fear & Surprise & Disgust & Total         \\ \hline
\cellcolor[HTML]{EFEFEF}\textbf{Best category-only} & 0.29    & 0.00  & 0.86 & 0.02  & 0.05 & 0.00     & 0.09    & \textbf{0.47} \\ \hline
\cellcolor[HTML]{EFEFEF}\textbf{Best combined} & 0.39    & 0.00  & 0.93 & 0.15  & 0.25 & 0.00     & 0.00    & \textbf{0.59} \\ \hline

\end{tabular}
\caption{Comparison of the best accuracies obtained on validation set by combined model and category-only model}
\label{table:compare_cate}
\end{table}

\begin{figure}[!htb]
\centering
\includegraphics[width = 1\hsize]{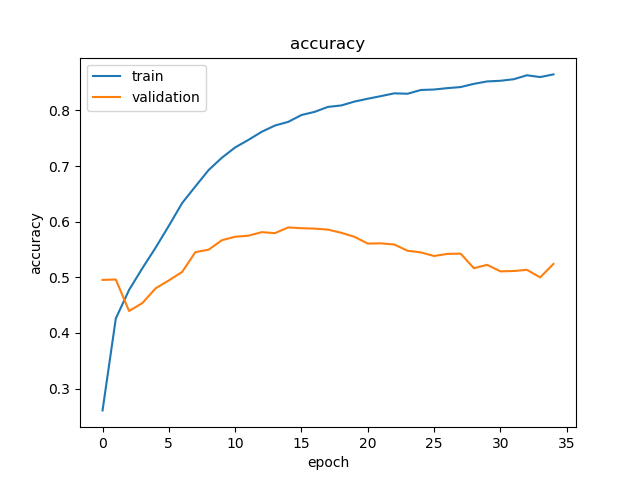}
\caption{Training history of the combined model achieving best results for categorical emotions}
\label{fig:acc_hist}
\end{figure}

\clearpage

\begin{table}[!htb]
\centering
\large
\begin{tabular}{|l|l|l|l|}
\hline
                                 & precision & recall & f1-score \\ \hline
\cellcolor[HTML]{EFEFEF}Neutral  & 0.57      & 0.39   & 0.46     \\ \hline
\cellcolor[HTML]{EFEFEF}Happy    & 0.00      & 0.00   & 0.00     \\ \hline
\cellcolor[HTML]{EFEFEF}Sad      & 0.64      & 0.93   & 0.76     \\ \hline
\cellcolor[HTML]{EFEFEF}Angry    & 0.15      & 0.15   & 0.15     \\ \hline
\cellcolor[HTML]{EFEFEF}Fear     & 0.26      & 0.25   & 0.26     \\ \hline
\cellcolor[HTML]{EFEFEF}Surprise & 0.00      & 0.00   & 0.00     \\ \hline
\cellcolor[HTML]{EFEFEF}Disgust  & 0.00      & 0.00   & 0.00     \\ \hline
\cellcolor[HTML]{EFEFEF}average  & 0.49      & 0.59   & 0.52     \\ \hline
\end{tabular}
\caption{Precision, accuracy, and f1-score of the VGG-19 combined architecture, which produces the best performance.}
\label{table:combined_cate_report}
\end{table}

The training history of the combined model obtaining the best accuracy is plotted in Figure~\ref{fig:acc_hist}. Both training and validation accuracy increase gradually. The models that achieved best validation loss were saved. \par

Detailed metrics including precision, recall and f1-score for the combined architecture that achieved the best performance can be found in Table~\ref{table:combined_cate_report}.

In summary, by constructing a combined model training both categorical and dimension labels together, the performance in categorical classification can be significantly improved. In addition, the effect of the imbalanced dataset is alleviated. \par

\section*{Dimensional Prediction Evaluation}

\begin{table}[htb]
\centering
\begin{tabular}{|c|c|c|c|c|}
\hline
                                             & \multicolumn{2}{c|}{CCC}      & \multicolumn{2}{c|}{MSE}      \\ \hline
                                             & Valence       & Arousal       & Valence       & Arousal       \\ \hline
\cellcolor[HTML]{EFEFEF}VGG16+LSTM+1FC          & 0.13          & 0.04          & 0.38          & 0.19          \\ \hline
\cellcolor[HTML]{EFEFEF}VGG16+LSTM+3FC          & 0.07          & 0.13          & 0.36          & 0.10          \\ \hline
\cellcolor[HTML]{EFEFEF}Xcep+LSTM+2FC           & -0.03          & 0.00          & 0.34          & 0.11          \\ \hline
\cellcolor[HTML]{EFEFEF}Xcep+LSTM+3FC           & 0.02          & 0.01          & 0.32          & 0.11          \\ \hline
\cellcolor[HTML]{EFEFEF}VGG19+LSTM+1FC          & 0.14          & 0.20          & 0.33          & 0.12          \\ \hline
\cellcolor[HTML]{EFEFEF}\textbf{VGG19+LSTM+2FC} & \textbf{0.14} & \textbf{0.20} & \textbf{0.33} & \textbf{0.10} \\ \hline
\end{tabular}
\caption{CCC and MSE evaluation of valence \& arousal predictions reached by the CNN-RNN architectures in combined models on validation set}
\label{table:combined_va}
\end{table}

Table~\ref{table:combined_va} summarises the evaluation of CCC and MSE values when using combined models on validation set. It can be seen that the best results have been achieved by VGG-19 + LSTM + 2 fully connected layers (128 units and 2 units). \par

Compared to the CCC values obtained in the dimensional-only models, there is no significant change that can be observed. Table~\ref{table:compare_va} summarised the comparison of best CCC and MSE evaluation of valence \& arousal predictions reached by the CNN-RNN architectures in the combined models and dimensional-only models on the validation set. It can be seen that the best CCCs achieved in dimensional representation only models are 0.16 and 0.17, compared to 0.14 and 0.20 in the combined models. The total MSE in the combined model increases by 0.03, whilst the total CCC decreases by 0.01.\par

\begin{table}[htb]
\centering
\begin{tabular}{|c|c|c|c|c|}
\hline
                                             & \multicolumn{2}{c|}{CCC}      & \multicolumn{2}{c|}{MSE}      \\ \hline
                                             & Valence       & Arousal       & Valence       & Arousal       \\ \hline
\cellcolor[HTML]{EFEFEF}\textbf{Best in Dimensional-only} & \textbf{0.16} & \textbf{0.17} & \textbf{0.31} & \textbf{0.09} \\ \hline
\cellcolor[HTML]{EFEFEF}\textbf{Best in Combined} & \textbf{0.14} & \textbf{0.20} & \textbf{0.33} & \textbf{0.10} \\ \hline
\end{tabular}
\caption{Comparison of best CCC and MSE evaluation of valence \& arousal predictions reached by the CNN-RNN architectures in combined models and dimensional-only models on validation set}
\label{table:compare_va}
\end{table}

\begin{figure}[htb]
\centering
\includegraphics[width = 0.825\hsize]{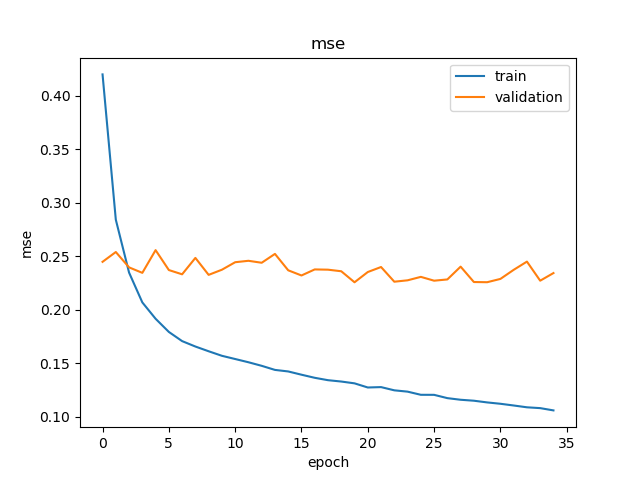}
\caption{Training history of the combined model achieving best results for dimensional emotion representation}
\label{fig:mse_hit}
\end{figure}

Figure~\ref{fig:mse_hit} plots the training history in the architecture with the best performance.  It can be observed that even in the architecture achieving best results, the loss on the validation set did not markedly decrease. The is mainly because the validation set is not representative enough. Sets have been split and optimised several times to achieve a high level of representation. However, due to the fact that there are 2 types of labels to be considered and there are only 55 videos, the labels of 2 representations are both independent and correlated, the space to adjust is relatively small. This issue will be discussed in future work and several solutions will be provided. \par

\clearpage

Another two observations can be made on all three models: \par
\begin{enumerate}
  \item Within all three types of architectures, the best performances have been achieved by the VGG-19 based CNN + RNN architectures. We infer from this that by increasing the depth of Convolutional layers, more useful image features can be extracted and benefit emotion recognition. 
  \item No evident increase can be obtained when simply increasing the layer of fully connected layers within each pre-trained networks. 
\end{enumerate}

In summary, by building the combined model and training the emotion category labels and dimensional valence and arousal values simultaneously, the results obtained for categorical prediction outperform the accuracies shown by the category-only model. Especially, the emotions which are minorities in the database start to be recognised and recalls are increased. \par

However, the CCC evaluations of valence and arousal do not demonstrate a marked increase in this method. The discussion and analysis of these phenomena are developed in the next section. \par

\clearpage

\section{Inner-Relationship of Emotion Category and Dimensional Representation Analysis}

In order to discover the reason for the increase in classification performance by training two affect labels together, and the relatively insignificant change in valence \& arousal evaluations, efforts were made on the analysis of the distribution of categorical emotions on valence and arousal space. Inner-relationship of two types of emotion representation is expected to be discovered, to explain the phenomenon displayed in the combined model.  \par

\begin{figure}[htb]
\centering
\includegraphics[width = 0.8\hsize]{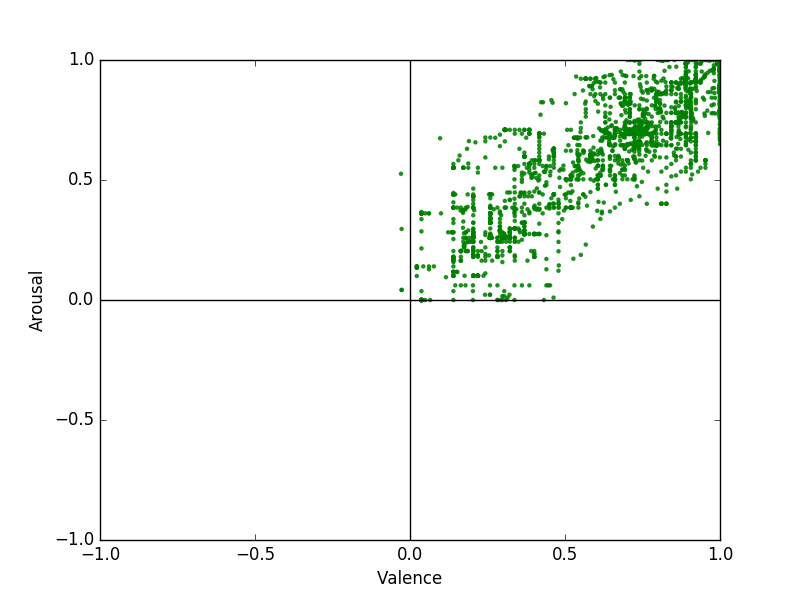}
\caption{Distribution of Happiness on Valence \& Arousal Dimension}
\label{fig:happiness}
\end{figure}

\begin{figure}[htb]
\centering
\includegraphics[width = 0.8\hsize]{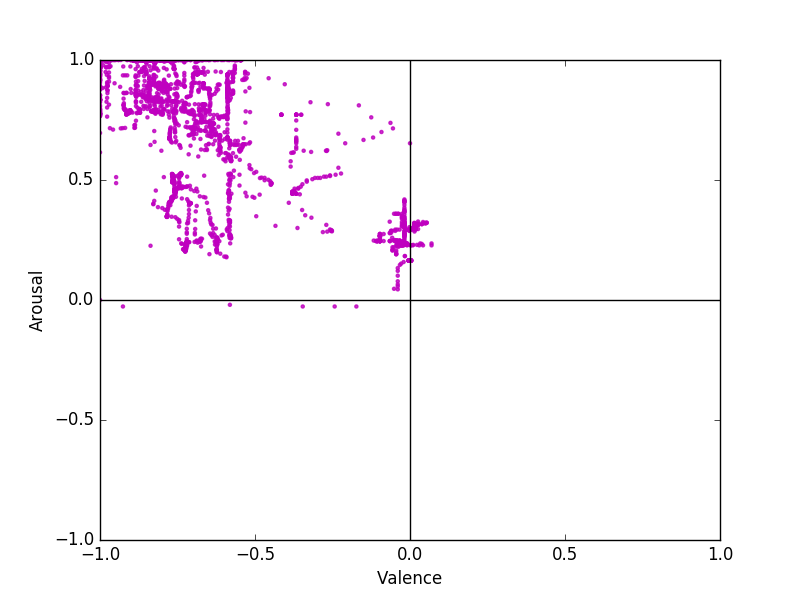}
\caption{Distribution of Fear on Valence \& Arousal Dimension}
\label{fig:fear}
\end{figure}

\begin{figure}[htb]
\centering
\includegraphics[width = 0.8\hsize]{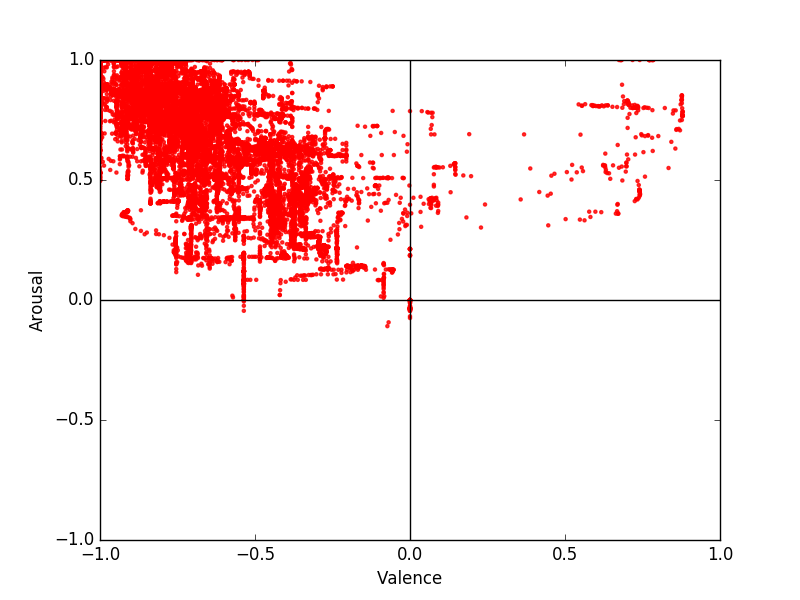}
\caption{Distribution of Sadness on Valence \& Arousal Dimension}
\label{fig:sadness}
\end{figure}

\begin{figure}[htb]
\centering
\includegraphics[width = 0.8\hsize]{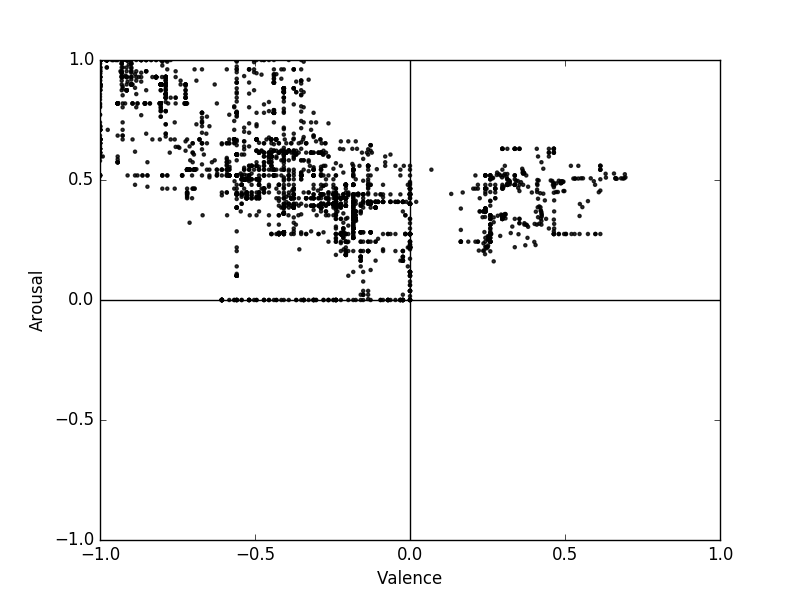}
\caption{Distribution of Disgust on Valence \& Arousal Dimension}
\label{fig:disgust}
\end{figure}

\begin{figure}[htb]
\centering
\includegraphics[width = 0.8\hsize]{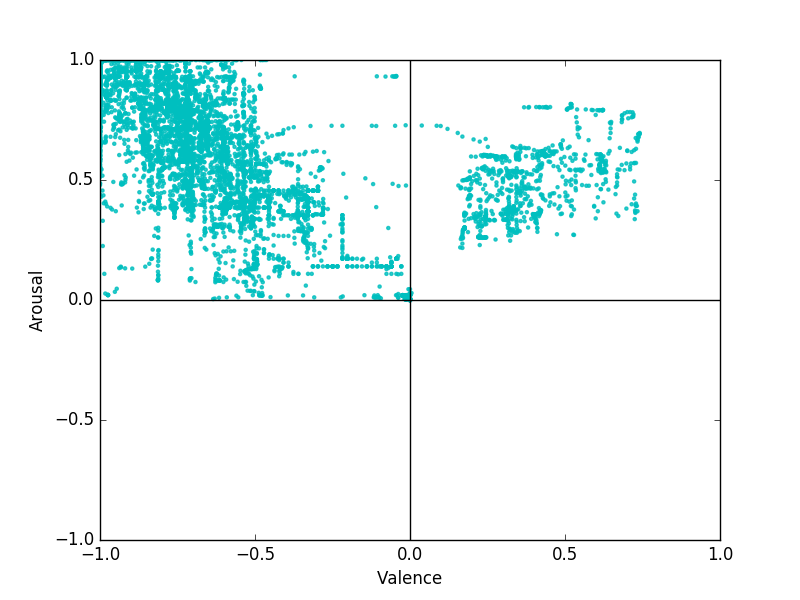}
\caption{Distribution of Angry on Valence \& Arousal Dimension}
\label{fig:angry}
\end{figure}

\begin{figure}[htb]
\centering
\includegraphics[width =0.8\hsize]{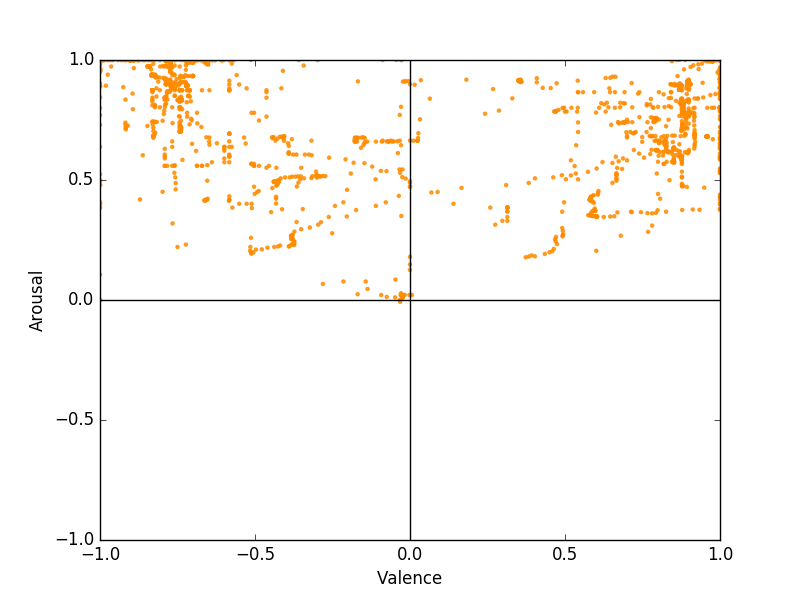}
\caption{Distribution of Surprise on Valence \& Arousal Dimension}
\label{fig:surprise}
\end{figure}

\begin{figure}[htb]
\centering
\includegraphics[width = 0.8\hsize]{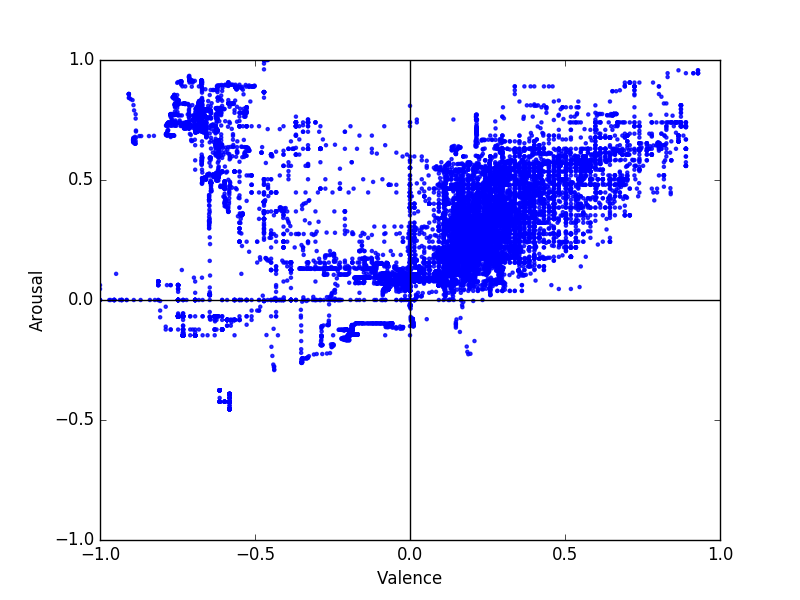}
\caption{Distribution of Neutral on Valence \& Arousal Dimension}
\label{fig:neutral}
\end{figure}

The distribution of the seven basic emotions on valence and arousal space are plotted in Figure~\ref{fig:happiness} to Figure~\ref{fig:neutral}. This is achieved by matching the two types of labels by the same video frame. 

It can be observed in Figure~\ref{fig:happiness} that Happiness category shows a strong relationship with Valence and Arousal. All most all the Happiness labels are located in first quadrant, where valence and arousal are larger than 0. 

Figure~\ref{fig:fear} shows that Fear illustrates a common characteristic of Happiness. The difference is most of Fears are located in the second quadrant, which means fear is a negative emotion. \par

The feature could be found in Figure~\ref{fig:sadness} that most of the sadness has a negative valance value. However, a small number of sadnesses are in the first quadrant. This means some sadness is classified as a positive emotion. This circumstance happens when the character in the video show emotions like "being moved". This distribution implies that a complex emotion annotation or labels combining both Valence \& Arousal could be a more accurate representation of some emotions.

Similarly, both Disgust and Angry cluster in the second quadrant. A small part of each emotion shows up in the valence-positive area, which inferred that small portion of Disgust and Angry could be positive in some situations. A more precise representation to characterize these kinds of emotions could be investigated in the future, and may lead to a more accurate emotion recognition.  

Particularly, Surprise demonstrates an equivalent possibility appearing in both positive and negative valence region. This makes sense because some Surprise in the database are more related to Fear, while some others are involved with Happiness, e.g. surprised by receiving an unexpected gift. 

In Figure~\ref{fig:neutral}, there is no specific pattern could be observed from distribution for Neutral. This is due to the fact that neutral is the most common emotion. Also, Neutral usually initiates the rest of six emotions, the various emotions initiated by neutral have an impact on neutral's valence \& arousal values. This leads to a wide distribution of neutral.  \par

\begin{figure}[htb]
\centering
\includegraphics[width = 0.9\hsize]{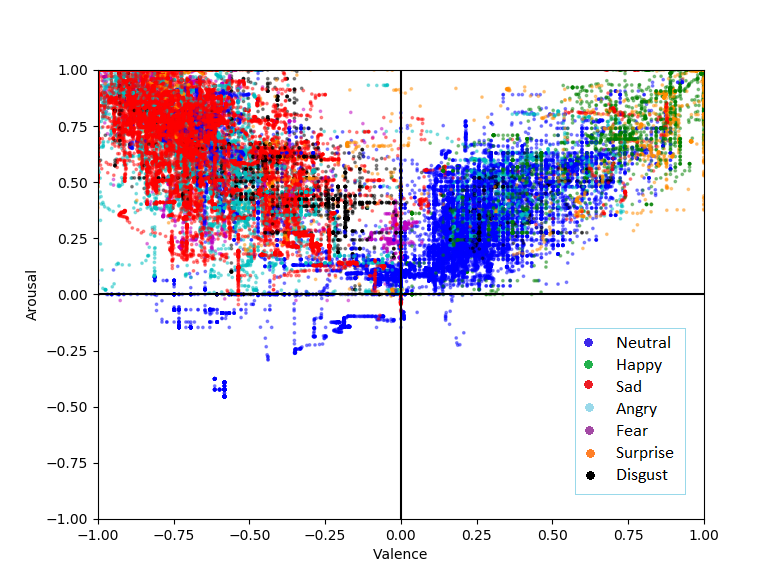}
\caption{Distribution of all emotions on Valence \& Arousal dimension}
\label{fig:all_emo}
\end{figure}

In conclusion, Happy and Fear shows a one-to-one-region mapping characteristic to the region of valence and arousal. Sadness, Disgust, Angry and Surprise demonstrates the one-to-two-region relationship when mapping to valence and arousal dimensions. In contrast, one specific valence and arousal pair can be mapped to at least 5 or 6 emotions. This feature is illustrated in Figure~\ref{fig:all_emo}, where all emotions label distributions are plotted on one space. 

This mapping difference might explain the performance increase of emotion category predictions in the combined model and no change in dimensional emotion evaluation in the combined models. 

\clearpage

\section{Inner-Relationship's Impact Analysis}

When we train the combined architectures with two annotations, the shared CNN and RNN part are expected to learn the features from both emotion representation. Therefore, when predicting the emotion in images, the features extracted by the CNN and RNN part are supposed to have extra information compared to single-function models. \par

For emotion category predictions, as described in the last section, a specific emotion can be mapped to one or two regions on the valence and arousal dimension. Hence, the extra information for the fully connected layer contains the features that map to one or two specific ranges of valence and arousal. This definite range is a clear signal for the neural network, and can benefit the neural network to make the prediction for emotion categories. \par

For example, when feeding an image which is supposed to be classified as angry to the category-only model, the neural network could make the prediction that the possibility of the image to be happy and angry are equal. In this situation, the possibility that the model can make the right prediction is 50\%. When feeding the same image to a combined model, the extra valence and arousal information will be extracted. In this case, the extra information can be interpreted as the valence and arousal are more likely to be negative. And the fully connected layer will receive this information, and predict the image to angry because of the distribution of angry. Because, compared to happy, angry is more likely to have a negative valence. Hence, the accuracy is improved. \par

On the contrary, the valence \& arousal output cannot benefit from this extra information extracted. As shown in Figure~\ref{fig:all_emo}, one pair of valence and arousal coordinates have the possibility to be all emotions. This extra information will be that this image can be more than five emotions. The message is not a clear signal, and does not have useful value for valence and arousal fully connected layers. It cannot help the layers to better predict the valence and arousal values. Even though the extra feature ideally implies only one emotion, as can be observed in Figure~\ref{fig:happiness} to Figure~\ref{fig:neutral}, one specific emotion can still spread in a range of areas. This information is still too general for the valence \& arousal fully connected layers to find to specific correct value. \par

In summary, the one-to-one and one-two mapping of the emotions to valence \& arousal ranges can help the combined model provide clearer features and make a better prediction. However, the one-to-many mapping of the valence \& arousal range to emotions cannot give the model specific information and the neural network cannot provide a better result. \par

\section{Interpretability Analysis}

Two techniques, t-Distributed Stochastic Neighbor Embedding and Activation Maximisation are carried out to discover the interpretability of the emotion recognition neural network. \par

\subsection*{t-Distributed Stochastic Neighbor Embedding}

t-Distributed Stochastic Neighbor Embedding \cite{maaten2008visualizing} is a machine learning technique that allows visualisation high-dimensional datapoints by reducing the dimensions to a two or three low dimensional map. t-Distributed Stochastic is an alteration of the original Stochastic Neighbor Embedding, which is easier to optimise and gives remarkably better visualisations \cite{vanDerMaaten2008}. \par

The idea of the Stochastic Neighbor Embedding is to convert the high-dimensional Euclidean distances between each datapoint into the conditional probabilities which can represent similarities. The definition of the similarity of datapoint $x_j$ to datapoint $x_i$ is the conditional probability $p_{j\mid i}$, which $x_i$ would pick $x_j$ as its neighbor if neighbors were picked in proportion to their probability density under a Gaussian centred at $x_i$ \cite{vanDerMaaten2008}:

\begin{equation}
p_{j\mid i} = \frac{\exp(-\lVert\mathbf{x}_i - \mathbf{x}_j\rVert^2 / 2\sigma_i^2)}{\sum_{k \neq i} \exp(-\lVert\mathbf{x}_i - \mathbf{x}_k\rVert^2 / 2\sigma_i^2)}
\end{equation}

where $\sigma_i$ is the variance of the Gaussian which is centred on datapoint $x_i$. 

\begin{figure}[htb]
\centering
\includegraphics[width = 1\hsize]{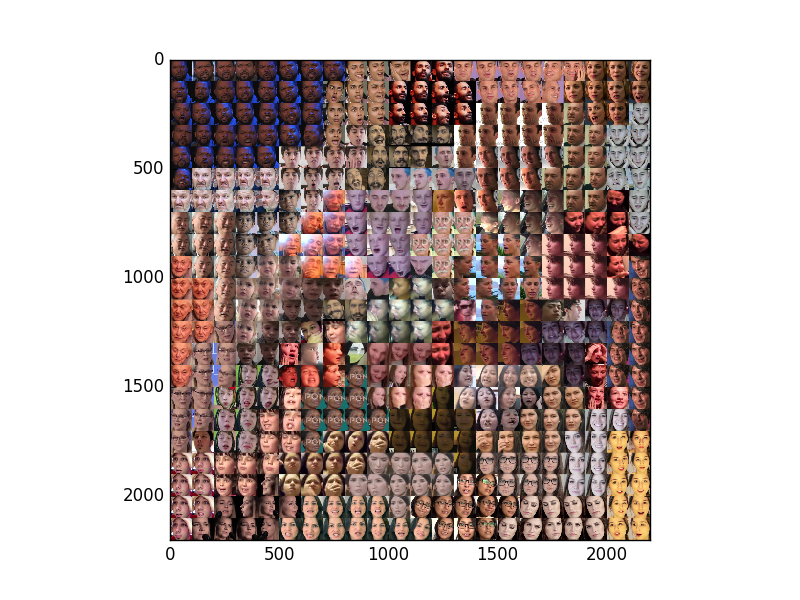}
\caption{Embedding neural network features into 2 dimensions.}
\label{fig:tsne1}
\end{figure}

Therefore, the closer the datapoints are, the $p_{j\mid i}$ is higher. And for the distantly separated datapoints, the $p_{j\mid i}$ will be close to zero. \par

t-Distributed Stochastic Neighbor Embedding optimised SNE by using a symmetrized SNE cost function with simpler gradients, and utilising the Student-t distribution instead of the Gaussian to compute the conditional probability between datapoints. These modifications enable a more successful visualisation on large datasets with less computational power. This technique can be used to illustrate how neural networks extract features in high-dimensional vectors, and visualise the similarity of the input data in 2-dimensional vectors. \par

t-SNE are used to explore how the neural network models interpret the emotions. Facial expression images were randomly selected from the database. t-SNE-Visualization \cite{tsne-visualization} package helps visualising the result. Facial images in database were fed into the CNN. High-dimensional features were extracted by convolutional layers and max pooling layers. t-SNE algorithm uses the high-dimensional features to embed the images into two dimensions. t-SNE calculates the conditional possibility that two features are similar, and visualise the distance between two features in 2 dimensions. Facial expressions being adjacent to each other in Figure~\ref{fig:tsne1} means that they are also close in the neural network representation space, hence, the neural network regards them as being alike. \par

It can be seen in Figure~\ref{fig:tsne1} that, the deep neural networks not only sees the images from the same person alike, but also sees the images with similar facial expressions or emotions as being similar. For example, happiness is closer to happiness and angry is closer to angry on the grid. In other words, t-SNE arranges images that have similar emotion features nearby after the embedding process. \par

\begin{figure}[htb]
\centering
\includegraphics[width = 1\hsize]{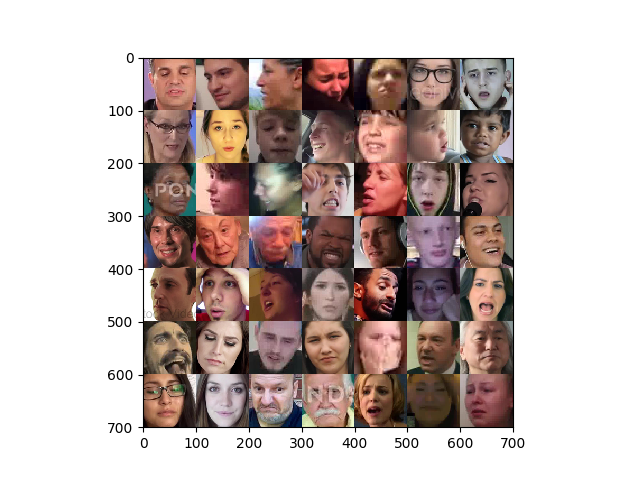}
\caption{Embedding neural network features into 2 dimensions}
\label{fig:tsne2}
\end{figure}

To further explore how the neural networks interpret the emotion and facial features in high dimensions, only one face was randomly selected from each video in the database. Feature extraction and t-SNE algorithm is applied to visualise the how neural network interpret facial features. \par 

In Figure~\ref{fig:tsne2} we can see that, even only fed one face from each video, the embedded 2 dimensions can still illustrate the pattern that similar facial expression are closer. \par

\begin{figure}[htb]
\centering
\includegraphics[width = 1\hsize]{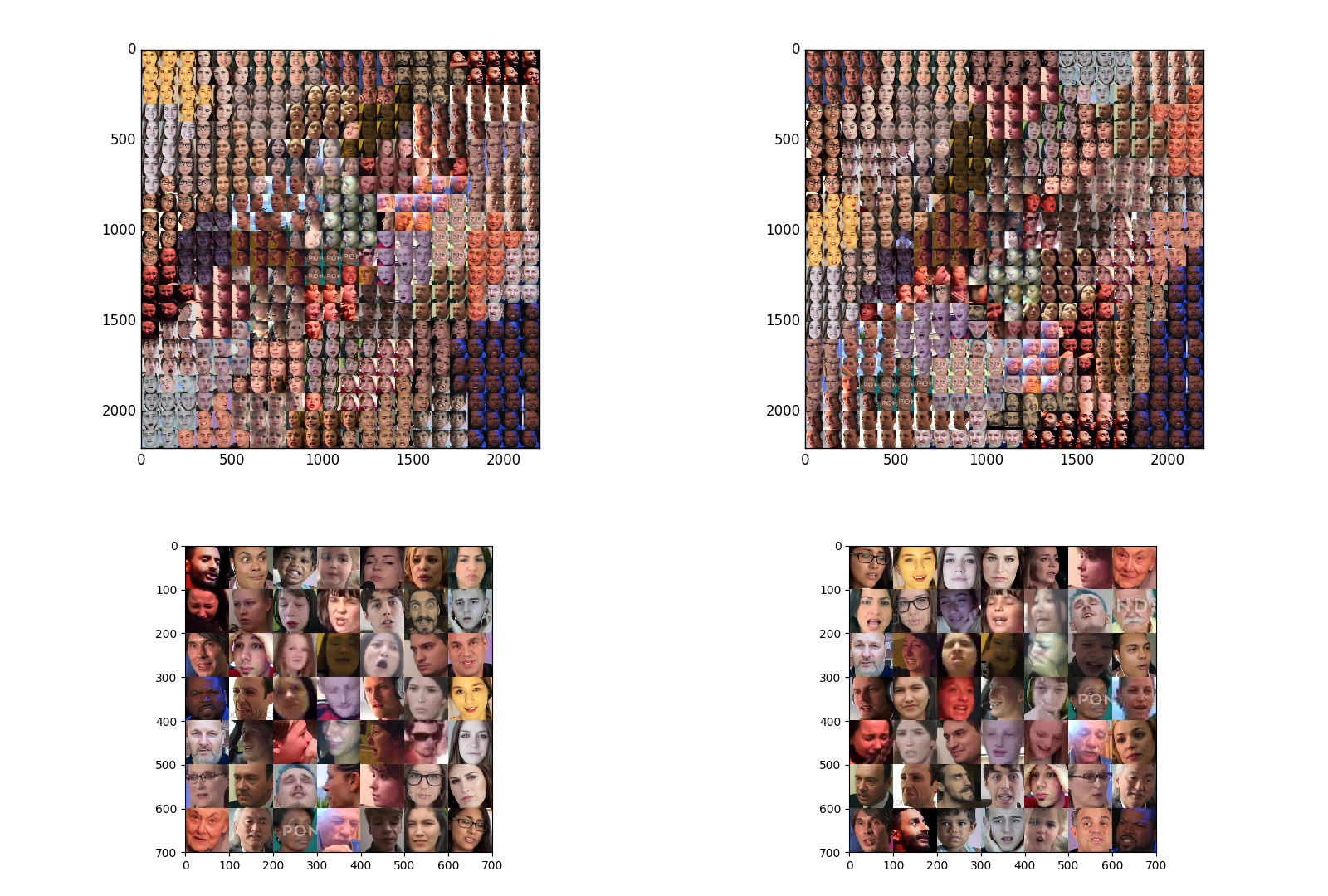}
\caption{More examples of t-SNE visualisation for the emotion recognition}
\label{fig:tsne_merge}
\end{figure}

Different pictures were randomly selected to test the interpretability of the architectures. More examples can be found in Figure~\ref{fig:tsne_merge}.

The result of t-SNE also implies that, there might exist a high-dimensional emotion representation E, which emotion categories and valence \& arousal are its respective mapping into one dimension and two dimensions. Mathematically, the mapping can be defined as:

\begin{equation}
emotion cateogry = f(E)
\end{equation}

\begin{equation}
valance \& arousal = g(E)
\end{equation}

The relationship of categorical representation and dimensional representation could be investigated through discovering this high-dimensional emotion representation. \par

\subsection*{Activation Maximisation}
Another technique attempted to discover the interpretability of the neural networks and what exactly neural networks learned is Activation Maximisation.\par

The Activation Maximisation was proposed by Dumitru Erhan et al. \cite{erhan2009visualizing}. The idea is to generate images that can maximumly activate the neurons for the layer. This allows the researcher to understand what data can activate particular layers, for example, CNN filters or categorical fully connected layers. Therefore, the researcher can better understand the function that is learned by the deep neural network. \par

To generate the Activation Maximisation images, the algorithm updates the synthesized photo by calculating the gradients to maximise the activation. Mathematically,

\begin{equation}
\frac{\partial Activation Maximisation Loss}{\partial Input Data}
\end{equation}

determines how to update the generated image. \par

\begin{figure}[htb]
\centering
\includegraphics[width = 0.7\hsize]{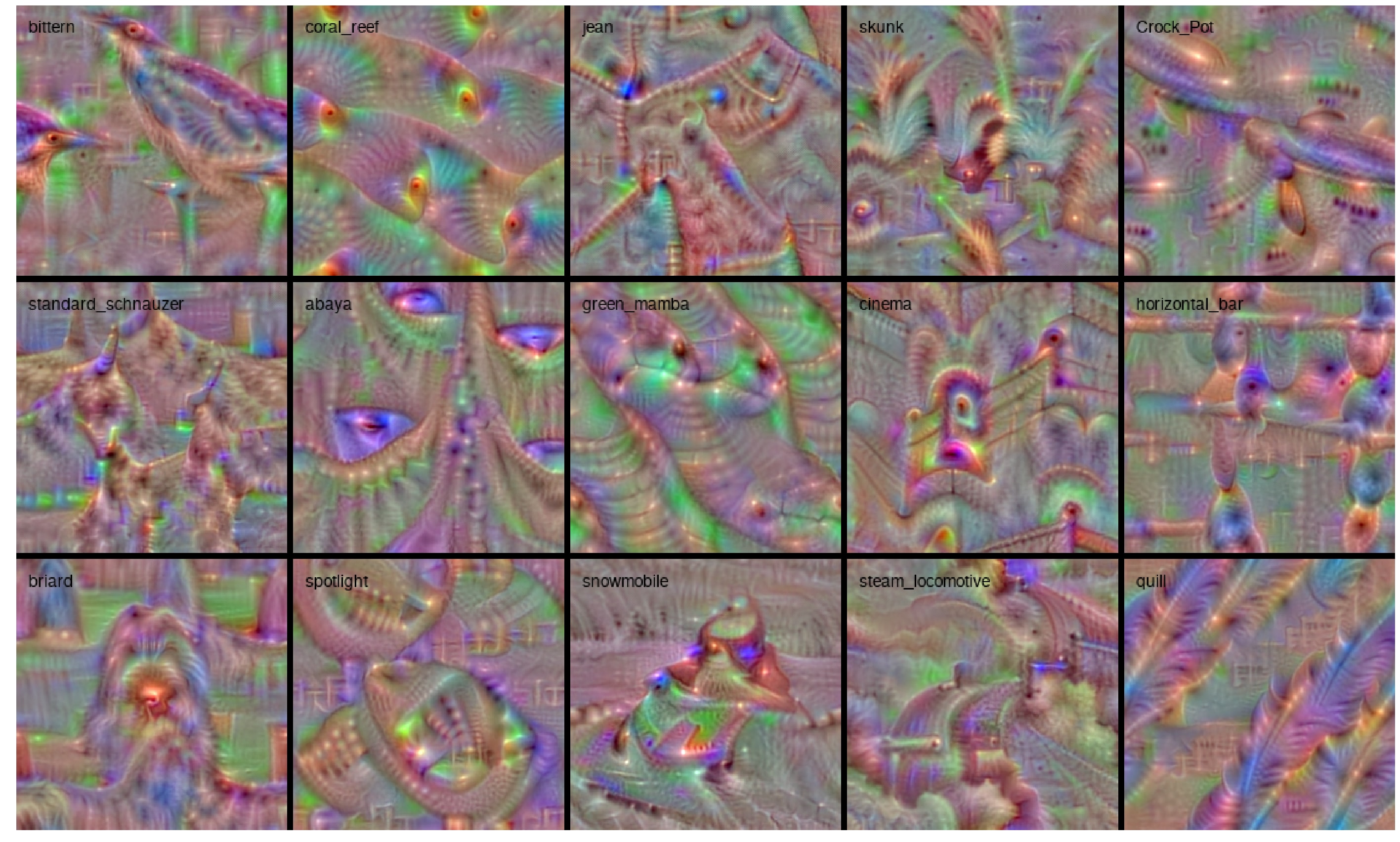}
\caption{Images that can maximum activate the last fully connected layers for some categories in VGG16 trained on ImageNet \cite{raghakotkerasvis}}
\label{fig:imagenet}
\end{figure}

Figure~\ref{fig:imagenet} visualises images that can maximumly activate the last fully connected layers for some categories in VGG16 trained on ImageNet database. It can be observed that some category features are humanly understandable. \par

Keras-vis \cite{raghakotkerasvis} toolkit is utilised to visualise the images that can activate the fully connected layers for seven emotions. 

\begin{figure}[htb]
\centering
\includegraphics[width = 0.6\hsize]{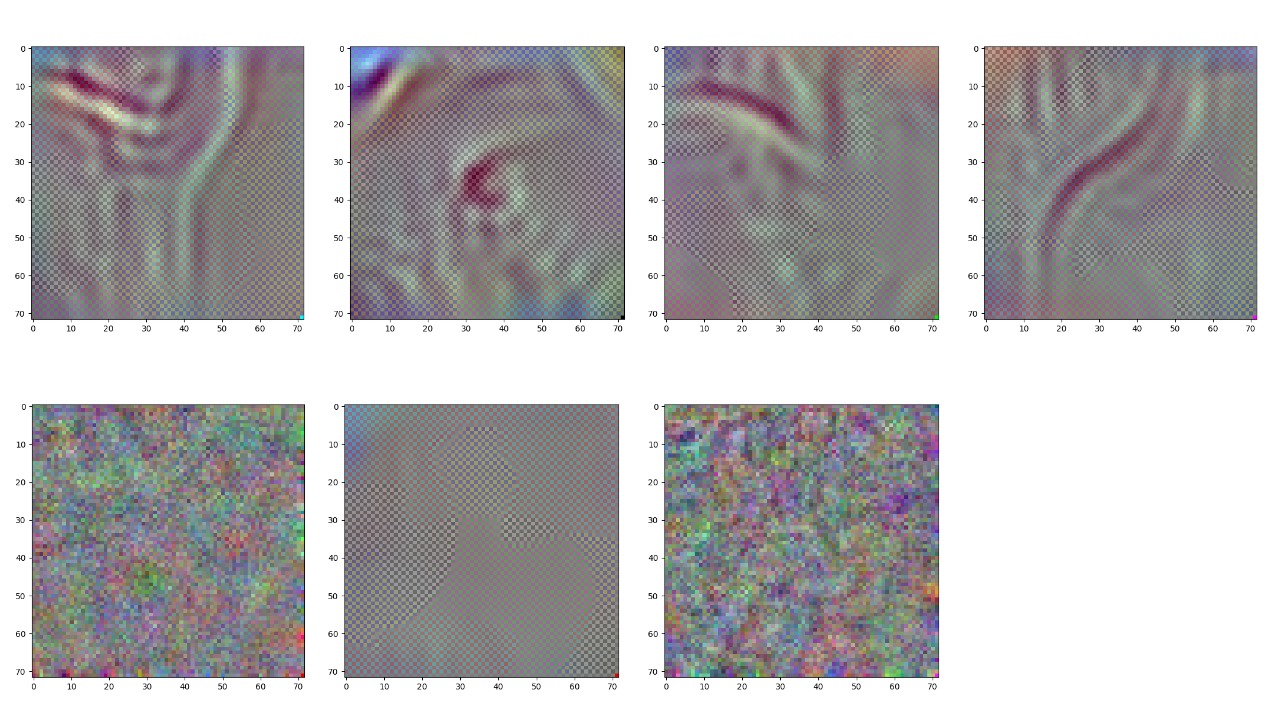}
\caption{Images that can maximumly activate the last fully connected layers for categorical emotion recognition model. (Left to right, top to buttom: Neutral, Happiness, Sadness, Angry, Fear, Surprise, Disgust)}
\label{fig:maxact}
\end{figure}

Figure ~\ref{fig:maxact} illustrates the images that can maximumly activate the last fully connected layers for categorical emotion recognition model. Although no specific features that the neural networks learned are particularly human-understandable, some categories such as neutral and sadness that can be better recognised in evaluation show more learned features. This technique can be applied to other models in the future to investigate the interpretability and reliability of emotion recognition models. \par





\chapter{Conclusion and Future Work}
This chapter summarises the contributions developed in this project. Several methods are provided for future work.  

\section{Contribution}

55 videos with 59 characters have been annotated with emotion categories. 107,640 frames of facial expressions with both emotion categories and valence \& arousal values have been made into the database for emotion recognition learning. \par

Different pre-trained CNN models and architectures have been attempted in building categorical-only models, dimensional-only models and combined models. Extensive experiments were carried out to optimise the performance and hyper-parameters. \par

The best classification accuracy increased by 25\% in the combined model compared to training with only emotion categories. While there is no obvious difference for performance of dimensional representation in different models. \par

The inner-relationship of emotion categories and valence \& arousal expression have been analysed. It is inferred that the extra information from mapping is the reason to explain the increased performance for categorical emotion prediction. \par

Two techniques, t-SNE and Activation Maximisation have been carried out to explore the interpretability of emotion recognition neural networks. Visualisations were used to interpret what the networks actually learned. \par

\section{Future work}

One limitation of this project is the database used is relatively small, and the emotions categories and valence \& arousal values are imbalanced. Emotion neutral and sadness occupies more than 80\% and the database. Negative arousal values are limited. \par

This leads to many problems, including:
\begin{enumerate}
  \item It is hard to adjust the training, validation and testing set to have the very close annotation distributions. 
  \item The predictions of different emotion categories vary a lot. 
  \item The evaluation of valence and arousal values on the validation set is not as good as on the training set. The model did not generalise well on the validation set. 
\end{enumerate}
Several methods can solve this problem. First, more videos could be included in the future. Especially for videos with emotion happiness, anger, fear, surprise and disgust. Also, more videos having negative arousals could be added.\par

Another way to alleviate those problems is to carry out data augmentation. For emotions and dimensional values that are under-represented, we can add duplicates to the databases. Many methods \cite{kollias8,kollias9} can generate duplicates of samples, for example, shifting, distorting, rotating and zooming in/out. \par

Generating synthetic samples is another good attempt to balance the database. Instead of adding duplicates, it is to generate artificial samples using more complicated algorithms. Generative Adversarial Networks can be utilised in the future for this project. \par

\subsection{Generative Adversarial Networks}
The Generative Adversarial Networks are deep neural network architectures that consist of two adversarial models, which is generative and discriminative respectively.  The generator takes in random noise \cite{raftopoulos2018beneficial} and gives 'fake' output. The discriminator will be fed with both real dataset and generated output from the generative model. The generator aims to produce output as 'real' as the true dataset. Meanwhile, the goal of the discriminator is to decide if its input belongs to the actual training dataset or not \cite{goodfellow2014generative} \cite{nicholson_gibson}. \par 

\begin{figure}[htb]
\centering
\includegraphics[width = 1\hsize]{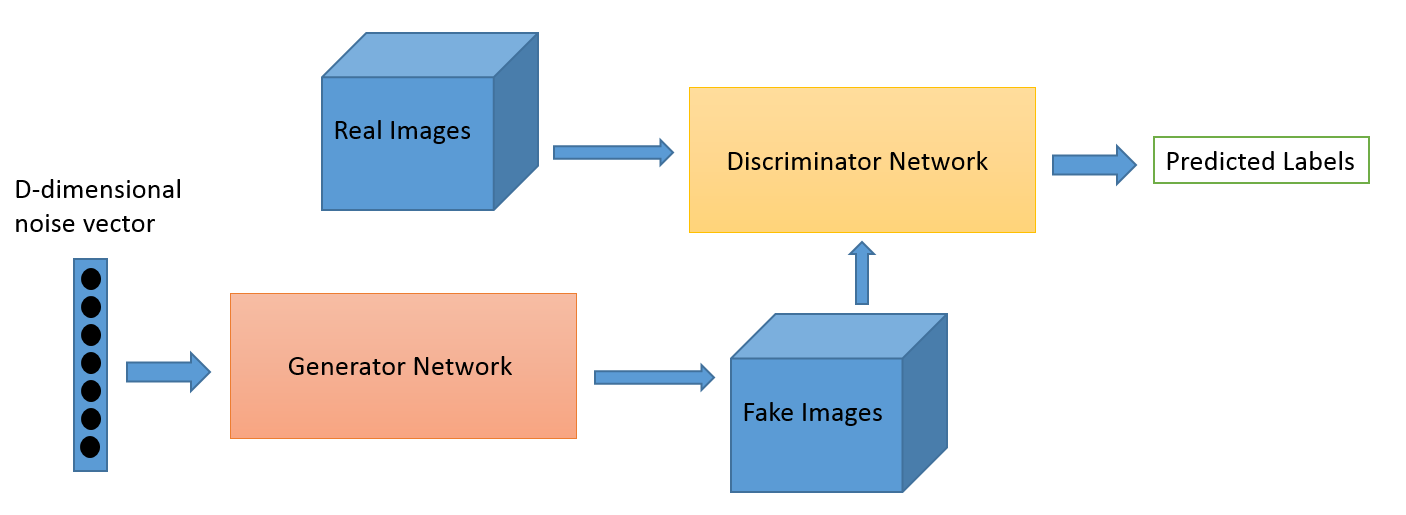}
\caption{A schematic of GAN. \cite{nicholson_gibson}}
\label{fig:gan}
\end{figure}

A graph of the principle of Generative Adversarial Networks can be seen in Figure ~\ref{fig:gan}.

The generative model can be utilised to generate images to fulfil the data augmentation. Also, the prediction of the emotion category from the discriminative model can be researched. Jost Tobias Springenberg reported that by applying semi-supervised learning using GANs, image classification result is competitive and the generated images are of high visual fidelity \cite{DBLP:journals/corr/Springenberg15}. 


\newpage

{\small
\bibliographystyle{ieee_fullname}
\bibliography{egbib}

\begin{thebibliography}{10}\itemsep=-1pt

\bibitem{lstm}
A beginner’s guide to recurrent networks and lstms.

\bibitem{cs231n}
Cs231n convolutional neural networks for visual recognition.

\bibitem{menpo14}
Joan {Alabort-i-Medina}, Epameinondas Antonakos, James Booth, Patrick Snape,
  and Stefanos Zafeiriou.
\newblock Menpo: A comprehensive platform for parametric image alignment and
  visual deformable models.
\newblock In {\em Proceedings of the ACM International Conference on
  Multimedia}, MM '14, pages 679--682, New York, NY, USA, 2014. ACM.

\bibitem{avrithis2000broadcast}
Yannis Avrithis, Nicolas Tsapatsoulis, and Stefanos Kollias.
\newblock Broadcast news parsing using visual cues: A robust face detection
  approach.
\newblock In {\em 2000 IEEE International Conference on Multimedia and Expo.
  ICME2000. Proceedings. Latest Advances in the Fast Changing World of
  Multimedia (Cat. No. 00TH8532)}, volume~3, pages 1469--1472. IEEE, 2000.

\bibitem{DBLP:journals/corr/Benitez-QuirozS17}
Carlos~Fabian Benitez{-}Quiroz, Ramprakash Srinivasan, Qianli Feng, Yan Wang,
  and Aleix~M. Mart{\'{\i}}nez.
\newblock Emotionet challenge: Recognition of facial expressions of emotion in
  the wild.
\newblock {\em CoRR}, abs/1703.01210, 2017.

\bibitem{buechel2018emotion}
Sven Buechel and Udo Hahn.
\newblock Emotion representation mapping for automatic lexicon construction
  (mostly) performs on human level.
\newblock {\em arXiv preprint arXiv:1806.08890}, 2018.

\bibitem{10.1007/978-0-387-74161-1_41}
George Caridakis, Ginevra Castellano, Loic Kessous, Amaryllis Raouzaiou, Lori
  Malatesta, Stelios Asteriadis, and Kostas Karpouzis.
\newblock Multimodal emotion recognition from expressive faces, body gestures
  and speech.
\newblock In Christos Boukis, Aristodemos Pnevmatikakis, and Lazaros
  Polymenakos, editors, {\em Artificial Intelligence and Innovations 2007: from
  Theory to Applications}, pages 375--388, Boston, MA, 2007. Springer US.

\bibitem{Chen:2017:MML:3133944.3133949}
Shizhe Chen, Qin Jin, Jinming Zhao, and Shuai Wang.
\newblock Multimodal multi-task learning for dimensional and continuous emotion
  recognition.
\newblock In {\em Proceedings of the 7th Annual Workshop on Audio/Visual
  Emotion Challenge}, AVEC '17, pages 19--26, New York, NY, USA, 2017. ACM.

\bibitem{DBLP:journals/corr/ChungGCB14}
Junyoung Chung, {\c{C}}aglar G{\"{u}}l{\c{c}}ehre, KyungHyun Cho, and Yoshua
  Bengio.
\newblock Empirical evaluation of gated recurrent neural networks on sequence
  modeling.
\newblock {\em CoRR}, abs/1412.3555, 2014.

\bibitem{Dhall:2017:IGE:3136755.3143004}
Abhinav Dhall, Roland Goecke, Shreya Ghosh, Jyoti Joshi, Jesse Hoey, and Tom
  Gedeon.
\newblock From individual to group-level emotion recognition: Emotiw 5.0.
\newblock In {\em Proceedings of the 19th ACM International Conference on
  Multimodal Interaction}, ICMI 2017, pages 524--528, New York, NY, USA, 2017.
  ACM.

\bibitem{du2015compound}
Shichuan Du and Aleix~M Martinez.
\newblock Compound facial expressions of emotion: from basic research to
  clinical applications.
\newblock {\em Dialogues in clinical neuroscience}, 17(4):443, 2015.

\bibitem{tsne-visualization}
e soroush.
\newblock tsne-visualization.

\bibitem{ekman1997face}
Paul Ekman and Erika~L Rosenberg.
\newblock {\em What the face reveals: Basic and applied studies of spontaneous
  expression using the Facial Action Coding System (FACS)}.
\newblock Oxford University Press, USA, 1997.

\bibitem{erhan2009visualizing}
Dumitru Erhan, Yoshua Bengio, Aaron Courville, and Pascal Vincent.
\newblock Visualizing higher-layer features of a deep network.
\newblock {\em University of Montreal}, 1341(3):1, 2009.

\bibitem{FoaEdnaB.1973Eith}
Edna~B. Foa.
\newblock Emotion in the human face: by paul ekman, wallace v. freesen and
  phoebe ellsworth pergamon press, new york, 1972. £3.50 (book review).
\newblock {\em Journal of Behavior Therapy and Experimental Psychiatry},
  4(1):87--88, 1973.

\bibitem{glimm2013using}
Birte Glimm, Yevgeny Kazakov, Ilianna Kollia, and Giorgos~B Stamou.
\newblock Using the tbox to optimise sparql queries.
\newblock {\em Description Logics}, 1014:181--196, 2013.

\bibitem{goodfellow2014generative}
Ian Goodfellow, Jean Pouget-Abadie, Mehdi Mirza, Bing Xu, David Warde-Farley,
  Sherjil Ozair, Aaron Courville, and Yoshua Bengio.
\newblock Generative adversarial nets.
\newblock In {\em Advances in neural information processing systems}, pages
  2672--2680, 2014.

\bibitem{goudelis2013exploring}
Georgios Goudelis, Konstantinos Karpouzis, and Stefanos Kollias.
\newblock Exploring trace transform for robust human action recognition.
\newblock {\em Pattern Recognition}, 46(12):3238--3248, 2013.

\bibitem{HAMANN2012458}
Stephan Hamann.
\newblock Mapping discrete and dimensional emotions onto the brain:
  controversies and consensus.
\newblock {\em Trends in Cognitive Sciences}, 16(9):458 -- 466, 2012.

\bibitem{horrocks2011answering}
Ilianna Kollia Birte Glimm~Ian Horrocks.
\newblock Answering queries over owl ontologies with sparql.
\newblock 2011.

\bibitem{kollia2011query}
Ilianna Kollia, Birte Glimm, and Ian Horrocks.
\newblock Query answering over sroiq knowledge bases with sparql.
\newblock In {\em Proceedings of the International Workshop on Description
  Logic}, 2011.

\bibitem{kollia2009interweaving}
Ilianna Kollia, Nikolaos Simou, Giorgos Stamou, and Andreas Stafylopatis.
\newblock Interweaving knowledge representation and adaptive neural networks.
\newblock In {\em Workshop on Inductive Reasoning and Machine Learning on the
  Semantic Web}, 2009.

\bibitem{kollias8}
Dimitrios Kollias, Shiyang Cheng, Maja Pantic, and Stefanos Zafeiriou.
\newblock Photorealistic facial synthesis in the dimensional affect space.
\newblock In {\em Proceedings of the European Conference on Computer Vision
  (ECCV)}, pages 0--0, 2018.

\bibitem{kollias9}
Dimitrios Kollias, Shiyang Cheng, Evangelos Ververas, Irene Kotsia, and
  Stefanos Zafeiriou.
\newblock Generating faces for affect analysis.
\newblock {\em arXiv preprint arXiv:1811.05027}, 2018.

\bibitem{kollias10}
Dimitris Kollias, George Marandianos, Amaryllis Raouzaiou, and Andreas-Georgios
  Stafylopatis.
\newblock Interweaving deep learning and semantic techniques for emotion
  analysis in human-machine interaction.
\newblock In {\em 2015 10th International Workshop on Semantic and Social Media
  Adaptation and Personalization (SMAP)}, pages 1--6. IEEE, 2015.

\bibitem{kollias2}
Dimitrios Kollias, Mihalis~A Nicolaou, Irene Kotsia, Guoying Zhao, and Stefanos
  Zafeiriou.
\newblock Recognition of affect in the wild using deep neural networks.
\newblock In {\em Proceedings of the IEEE Conference on Computer Vision and
  Pattern Recognition Workshops}, pages 26--33, 2017.

\bibitem{kollias11}
Dimitrios Kollias, Athanasios Tagaris, and Andreas Stafylopatis.
\newblock On line emotion detection using retrainable deep neural networks.
\newblock In {\em 2016 IEEE Symposium Series on Computational Intelligence
  (SSCI)}, pages 1--8. IEEE, 2016.

\bibitem{kollias13}
Dimitrios Kollias, Athanasios Tagaris, Andreas Stafylopatis, Stefanos Kollias,
  and Georgios Tagaris.
\newblock Deep neural architectures for prediction in healthcare.
\newblock {\em Complex \& Intelligent Systems}, 4(2):119--131, 2018.

\bibitem{kollias3}
Dimitrios Kollias, Panagiotis Tzirakis, Mihalis~A Nicolaou, Athanasios
  Papaioannou, Guoying Zhao, Bj{\"o}rn Schuller, Irene Kotsia, and Stefanos
  Zafeiriou.
\newblock Deep affect prediction in-the-wild: Aff-wild database and challenge,
  deep architectures, and beyond.
\newblock {\em International Journal of Computer Vision}, 127(6-7):907--929,
  2019.

\bibitem{kollias12}
Dimitrios Kollias, Miao Yu, Athanasios Tagaris, Georgios Leontidis, Andreas
  Stafylopatis, and Stefanos Kollias.
\newblock Adaptation and contextualization of deep neural network models.
\newblock In {\em 2017 IEEE Symposium Series on Computational Intelligence
  (SSCI)}, pages 1--8. IEEE, 2017.

\bibitem{kollias4}
Dimitrios Kollias and Stefanos Zafeiriou.
\newblock Aff-wild2: Extending the aff-wild database for affect recognition.
\newblock {\em arXiv preprint arXiv:1811.07770}, 2018.

\bibitem{kollias7}
Dimitrios Kollias and Stefanos Zafeiriou.
\newblock A multi-component cnn-rnn approach for dimensional emotion
  recognition in-the-wild.
\newblock {\em arXiv preprint arXiv:1805.01452}, 2018.

\bibitem{kollias5}
Dimitrios Kollias and Stefanos Zafeiriou.
\newblock A multi-task learning \& generation framework: Valence-arousal,
  action units \& primary expressions.
\newblock {\em arXiv preprint arXiv:1811.07771}, 2018.

\bibitem{kollias6}
Dimitrios Kollias and Stefanos Zafeiriou.
\newblock Training deep neural networks with different datasets in-the-wild:
  The emotion recognition paradigm.
\newblock In {\em 2018 International Joint Conference on Neural Networks
  (IJCNN)}, pages 1--8. IEEE, 2018.

\bibitem{kollias14}
Dimitrios Kollias and Stefanos Zafeiriou.
\newblock Exploiting multi-cnn features in cnn-rnn based dimensional emotion
  recognition on the omg in-the-wild dataset.
\newblock {\em arXiv preprint arXiv:1910.01417}, 2019.

\bibitem{kollias15}
Dimitrios Kollias and Stefanos Zafeiriou.
\newblock Expression, affect, action unit recognition: Aff-wild2, multi-task
  learning and arcface.
\newblock {\em arXiv preprint arXiv:1910.04855}, 2019.

\bibitem{raghakotkerasvis}
Raghavendra Kotikalapudi and contributors.
\newblock keras-vis, 2017.

\bibitem{KrakovnaViktoriya2016BIMF}
Viktoriya Krakovna.
\newblock Building interpretable models: From bayesian networks to neural
  networks, January 2016.

\bibitem{5537907}
Y. LeCun, K. Kavukcuoglu, and C. Farabet.
\newblock Convolutional networks and applications in vision.
\newblock In {\em Proceedings of 2010 IEEE International Symposium on Circuits
  and Systems}, pages 253--256, May 2010.

\bibitem{doi:10.1093/cercor/bhk024}
PA Lewis, HD Critchley, P Rotshtein, and RJ Dolan.
\newblock Neural correlates of processing valence and arousal in affective
  words.
\newblock {\em Cerebral Cortex}, 17(3):742--748, 2007.

\bibitem{maaten2008visualizing}
Laurens van~der Maaten and Geoffrey Hinton.
\newblock Visualizing data using t-sne.
\newblock {\em Journal of machine learning research}, 9(Nov):2579--2605, 2008.

\bibitem{10.1007/978-3-319-10593-2_47}
Markus Mathias, Rodrigo Benenson, Marco Pedersoli, and Luc Van~Gool.
\newblock Face detection without bells and whistles.
\newblock In David Fleet, Tomas Pajdla, Bernt Schiele, and Tinne Tuytelaars,
  editors, {\em Computer Vision -- ECCV 2014}, pages 720--735, Cham, 2014.
  Springer International Publishing.

\bibitem{nicholson_gibson}
Chris~V. Nicholson and Adam Gibson.
\newblock Gan: A beginner's guide to generative adversarial networks.

\bibitem{DBLP:journals/corr/PapernotMGJCS16}
Nicolas Papernot, Patrick~D. McDaniel, Ian~J. Goodfellow, Somesh Jha, Z.~Berkay
  Celik, and Ananthram Swami.
\newblock Practical black-box attacks against deep learning systems using
  adversarial examples.
\newblock {\em CoRR}, abs/1602.02697, 2016.

\bibitem{raftopoulos2018beneficial}
Konstantinos~A Raftopoulos, Stefanos~D Kollias, Dionysios~D Sourlas, and Marin
  Ferecatu.
\newblock On the beneficial effect of noise in vertex localization.
\newblock {\em International Journal of Computer Vision}, 126(1):111--139,
  2018.

\bibitem{raghu2017svcca}
Maithra Raghu, Justin Gilmer, Jason Yosinski, and Jascha Sohl-Dickstein.
\newblock Svcca: Singular vector canonical correlation analysis for deep
  learning dynamics and interpretability.
\newblock In {\em Advances in Neural Information Processing Systems}, pages
  6076--6085, 2017.

\bibitem{nott45489}
Fabien Ringeval, Bj{\"o}rn Schuller, Michel Valstar, Jonathan Gratch, Roddy
  Cowie, Stefan Scherer, Sharon Mozgai, Nicholas Cummins, Maximilian Schmitt,
  and Maja Pantic.
\newblock Avec 2017--real-life depression, and affect recognition workshop and
  challenge.
\newblock In {\em 7th Audio/Visual Emotion Challenge and Workshop}, pages 3--9,
  October 2017.
\newblock Published in: Proceedings of the 7th International Workshop on
  Audio/Visual Emotion Challenge. New York : ACM, 2017, p. 3-9.
  doi:10.1145/3133944.3133953.

\bibitem{ringeval2017avec}
Fabien Ringeval, Bj{\"o}rn Schuller, Michel Valstar, Jonathan Gratch, Roddy
  Cowie, Stefan Scherer, Sharon Mozgai, Nicholas Cummins, Maximilian Schmitt,
  and Maja Pantic.
\newblock Avec 2017: Real-life depression, and affect recognition workshop and
  challenge.
\newblock In {\em Proceedings of the 7th Annual Workshop on Audio/Visual
  Emotion Challenge}, pages 3--9. ACM, 2017.

\bibitem{43146}
D. Sculley, Gary Holt, Daniel Golovin, Eugene Davydov, Todd Phillips, Dietmar
  Ebner, Vinay Chaudhary, and Michael Young.
\newblock Machine learning: The high interest credit card of technical debt.
\newblock In {\em SE4ML: Software Engineering for Machine Learning (NIPS 2014
  Workshop)}, 2014.

\bibitem{simonyan2013deep}
Karen Simonyan, Andrea Vedaldi, and Andrew Zisserman.
\newblock Deep inside convolutional networks: Visualising image classification
  models and saliency maps.
\newblock {\em arXiv preprint arXiv:1312.6034}, 2013.

\bibitem{DBLP:journals/corr/SimonyanZ14a}
Karen Simonyan and Andrew Zisserman.
\newblock Very deep convolutional networks for large-scale image recognition.
\newblock {\em CoRR}, abs/1409.1556, 2014.

\bibitem{simou2008image}
Nikos Simou, Th Athanasiadis, Giorgos Stoilos, and Stefanos Kollias.
\newblock Image indexing and retrieval using expressive fuzzy description
  logics.
\newblock {\em Signal, Image and Video Processing}, 2(4):321--335, 2008.

\bibitem{simou2007fire}
Nikolaos Simou and Stefanos Kollias.
\newblock Fire: A fuzzy reasoning engine for impecise knowledge.
\newblock In {\em K-Space PhD Students Workshop, Berlin, Germany}, volume~14.
  Citeseer, 2007.

\bibitem{spark}
Cambridge Spark.
\newblock Deep learning for complete beginners: convolutional neural networks
  with keras.

\bibitem{DBLP:journals/corr/Springenberg15}
Jost~Tobias Springenberg.
\newblock Unsupervised and semi-supervised learning with categorical generative
  adversarial networks.
\newblock {\em CoRR}, abs/1511.06390, 2015.

\bibitem{tagaris1}
Athanasios Tagaris, Dimitrios Kollias, and Andreas Stafylopatis.
\newblock Assessment of parkinson’s disease based on deep neural networks.
\newblock In {\em International Conference on Engineering Applications of
  Neural Networks}, pages 391--403. Springer, 2017.

\bibitem{tagaris2}
Athanasios Tagaris, Dimitrios Kollias, Andreas Stafylopatis, Georgios Tagaris,
  and Stefanos Kollias.
\newblock Machine learning for neurodegenerative disorder diagnosis—survey of
  practices and launch of benchmark dataset.
\newblock {\em International Journal on Artificial Intelligence Tools},
  27(03):1850011, 2018.

\bibitem{Tan:2017:GER:3136755.3143008}
Lianzhi Tan, Kaipeng Zhang, Kai Wang, Xiaoxing Zeng, Xiaojiang Peng, and Yu
  Qiao.
\newblock Group emotion recognition with individual facial emotion cnns and
  global image based cnns.
\newblock In {\em Proceedings of the 19th ACM International Conference on
  Multimodal Interaction}, ICMI 2017, pages 549--552, New York, NY, USA, 2017.
  ACM.

\bibitem{6020812}
M.~F. Valstar and M. Pantic.
\newblock Fully automatic recognition of the temporal phases of facial actions.
\newblock {\em IEEE Transactions on Systems, Man, and Cybernetics, Part B
  (Cybernetics)}, 42(1):28--43, Feb 2012.

\bibitem{vanDerMaaten2008}
Laurens van~der Maaten and Geoffrey Hinton.
\newblock Visualizing data using {t-SNE}.
\newblock {\em Journal of Machine Learning Research}, 9:2579--2605, 2008.

\bibitem{7292443}
A.~A. Varghese, J.~P. Cherian, and J.~J. Kizhakkethottam.
\newblock Overview on emotion recognition system.
\newblock In {\em 2015 International Conference on Soft-Computing and Networks
  Security (ICSNS)}, pages 1--5, Feb 2015.

\bibitem{article1}
yi-hsuan Yang and Homer Chen.
\newblock Machine recognition of music emotion: A review.
\newblock 3, 05 2012.

\bibitem{kollias1}
Stefanos Zafeiriou, Dimitrios Kollias, Mihalis~A Nicolaou, Athanasios
  Papaioannou, Guoying Zhao, and Irene Kotsia.
\newblock Aff-wild: Valence and arousal'in-the-wild'challenge.
\newblock In {\em Proceedings of the IEEE Conference on Computer Vision and
  Pattern Recognition Workshops}, pages 34--41, 2017.

\end{thebibliography}
}
\newpage

\newpage
\appendix

\chapter{Ethics Checklist}
{
\renewcommand*{\arraystretch}{1.3}
\begin{longtable}{ |p{13.2cm}|p{0.6cm}|p{0.6cm}| }
\hline
 & \bf Yes & \bf No \\
\hline

\multicolumn{3}{|l|}{\cellcolor{green!25}\bf Section 1: HUMAN EMBRYOS/FOETUSES} \\
\hline

Does your project involve Human Embryonic Stem Cells? & & \checkmark\\
\hline

Does your project involve the use of human embryos? & & \checkmark\\
\hline

Does your project involve the use of human foetal tissues / cells? & & \checkmark\\
\hline

\multicolumn{3}{|l|}{\cellcolor{green!25}\bf Section 2: HUMANS} \\
\hline

Does your project involve human participants? & \checkmark & \\
\hline

\multicolumn{3}{|l|}{\cellcolor{green!25}\bf Section 3: HUMAN CELLS / TISSUES} \\
\hline

Does your project involve human cells or tissues? (Other than from “Human Embryos/Foetuses” i.e. Section 1)? & & \checkmark\\
\hline

\multicolumn{3}{|l|}{\cellcolor{green!25}\bf Section 4: PROTECTION OF PERSONAL DATA} \\
\hline

Does your project involve personal data collection and/or processing? & \checkmark & \\
\hline

Does it involve the collection and/or processing of sensitive personal data (e.g. health, sexual lifestyle, ethnicity, political opinion, religious or philosophical conviction)? & & \checkmark\\
\hline

Does it involve processing of genetic information? & & \checkmark \\
\hline

Does it involve tracking or observation of participants? It should be noted that this issue is not limited to surveillance or localization data. It also applies to Wan data such as IP address, MACs, cookies etc. & \checkmark& \\
\hline

Does your project involve further processing of previously collected personal data (secondary use)? For example Does your project involve merging existing data sets? & & \checkmark \\
\hline

\multicolumn{3}{|l|}{\cellcolor{green!25}\bf Section 5: ANIMALS} \\
\hline

Does your project involve animals? &  & \checkmark  \\
\hline

\multicolumn{3}{|l|}{\cellcolor{green!25}\bf Section 6: DEVELOPING COUNTRIES} \\
\hline

Does your project involve developing countries? & & \checkmark \\
\hline

If your project involves low and/or lower-middle income countries, are any benefit-sharing actions planned? & & \checkmark  \\
\hline

Could the situation in the country put the individuals taking part in the project at risk? & & \checkmark  \\
\hline

\multicolumn{3}{|l|}{\cellcolor{green!25}\bf Section 7: ENVIRONMENTAL PROTECTION AND SAFETY} \\
\hline

Does your project involve the use of elements that may cause harm to the environment, animals or plants? & &  \checkmark  \\
\hline

Does your project deal with endangered fauna and/or flora /protected areas? & &  \checkmark   \\
\hline

Does your project involve the use of elements that may cause harm to humans, including project staff? & &  \checkmark   \\
\hline

Does your project involve other harmful materials or equipment, e.g. high-powered laser systems? & &  \checkmark   \\
\hline

\multicolumn{3}{|l|}{\cellcolor{green!25}\bf Section 8: DUAL USE} \\
\hline

Does your project have the potential for military applications? & &  \checkmark   \\
\hline

Does your project have an exclusive civilian application focus? & &  \checkmark   \\
\hline

Will your project use or produce goods or information that will require export licenses in accordance with legislation on dual use items? & &  \checkmark   \\
\hline

Does your project affect current standards in military ethics – e.g., global ban on weapons of mass destruction, issues of proportionality, discrimination of combatants and accountability in drone and autonomous robotics developments, incendiary or laser weapons? & &  \checkmark   \\
\hline

\multicolumn{3}{|l|}{\cellcolor{green!25}\bf Section 9: MISUSE} \\
\hline

Does your project have the potential for malevolent/criminal/terrorist abuse? & &  \checkmark   \\
\hline

Does your project involve information on/or the use of biological-, chemical-, nuclear/radiological-security sensitive materials and explosives, and means of their delivery? & &  \checkmark   \\
\hline

Does your project involve the development of technologies or the creation of information that could have severe negative impacts on human rights standards (e.g. privacy, stigmatization, discrimination), if misapplied? & &  \checkmark   \\
\hline

Does your project have the potential for terrorist or criminal abuse e.g. infrastructural vulnerability studies, cybersecurity related project? & &   \checkmark  \\
\hline

\multicolumn{3}{|l|}{\cellcolor{green!25}\bf Section 10: LEGAL ISSUES} \\
\hline

Will your project use or produce software for which there are copyright licensing implications? & &  \checkmark   \\
\hline

Will your project use or produce goods or information for which there are data protection, or other legal implications? & &   \checkmark  \\
\hline

\multicolumn{3}{|l|}{\cellcolor{green!25}\bf Section 11: OTHER ETHICS ISSUES} \\
\hline

Are there any other ethics issues that should be taken into consideration? & &  \checkmark   \\
\hline

\end{longtable}
}

\chapter{Ethical and Professional Considerations}
This project is about emotion recognition. It does not involve human embryos/cells and animals. Developing countries are not involved. It does not include elements that may cause harm to the environment. It has not been found that this project has the potential for dual use/military application and misuse issues. \par 

Areas that may have ethical issues are the video data collection and the involved human beings. For the videos researched in this project, the collection has been conducted under the scrutiny and approval of the Imperial College Ethical Committee (ICREC). Most of the videos were under Creative Commons License (CCL). For some videos which were not under Creative Commons License (CCL), producer or the person who is in the video have been contacted. Permissions to use the video for this research have been granted. There are not other legal and ethical issues. \par

\chapter{Evaluation}

\section{Evaluation of the Models on Training Set}
\subsection*{Category-only Models}
\begin{table}[!htb]
\centering
\begin{tabular}{|c|c|c|c|c|c|c|c|c|}
\hline
                                           & \multicolumn{8}{c|}{Accuracy}                                              \\ \hline
                                           & Neutral & Happy & Sad  & Angry & Fear & Surprise & Disgust & Total         \\ \hline
\cellcolor[HTML]{EFEFEF}VGG16+LSTM+1FC          & 0.52    & 0.00  & 0.98 & 0.00  & 0.00 & 0.00     & 0.00    & 0.52          \\ \hline
\cellcolor[HTML]{EFEFEF}VGG16+LSTM+2FC          & 0.90    & 0.00  & 0.97 & 0.55  & 0.56 & 0.00     & 0.71    & 0.78          \\ \hline
\cellcolor[HTML]{EFEFEF}Xcep+LSTM+1FC           & 0.44    & 0.00  & 0.77 & 0.01  & 0.00 & 0.00     & 0.00    & 0.42          \\ \hline
\cellcolor[HTML]{EFEFEF}Xcep+LSTM+2FC           & 0.58    & 0.00  & 0.70 & 0.06  & 0.00 & 0.00     & 0.00    & 0.44          \\ \hline
\cellcolor[HTML]{EFEFEF}VGG19+LSTM+1FC & 0.43    & 0.00  & 0.97 & 0.01  & 0.00 & 0.00     & 0.00    & 0.49 \\ \hline
\cellcolor[HTML]{EFEFEF}VGG19+LSTM+2FC          & 0.78    & 0.00  & 0.78 & 0.00  & 0.00 & 0.00     & 0.00    & 0.53          \\ \hline
\end{tabular}
\caption{Accuracies obtained on training set by various CNN-RNN architectures for categorical-only models}
\label{table:cate_res_train}
\end{table}

\clearpage

\subsection*{Dimension-only Models}
\begin{table}[!htb]
\centering
\begin{tabular}{|c|c|c|c|c|}
\hline
                                    & \multicolumn{2}{c|}{CCC}      & \multicolumn{2}{c|}{MSE}      \\ \hline
                                    & Valence       & Arousal       & Valence       & Arousal       \\ \hline
\cellcolor[HTML]{EFEFEF}VGG16+LSTM+1FC & 0.86          & 0.76       & 0.07        & 0.04        \\ \hline
\cellcolor[HTML]{EFEFEF}VGG16+LSTM+2FC & 0.83        & 0.66        & 0.08        & 0.05        \\ \hline
\cellcolor[HTML]{EFEFEF}Xcep+LSTM+1FC  & 0.03          & 0.05          & 0.31          & 0.11          \\ \hline
\cellcolor[HTML]{EFEFEF}Xcep+LSTM+2FC  & 0.01          & 0.01          & 0.29          & 0.10          \\ \hline
\cellcolor[HTML]{EFEFEF}VGG19+LSTM+1FC & 0.73        & 0.60       & 0.11        & 0.06        \\ \hline
\cellcolor[HTML]{EFEFEF}VGG19+LSTM +2FC & 0.87 & 0.72 & 0.07 & 0.04 \\ \hline
\end{tabular}
\caption{CCC and MSE evaluation of valence \& arousal predictions reached by the CNN-RNN architectures in dimensional-only models on training set}
\label{table:va_train}
\end{table}

\clearpage

\subsection*{Combined Models Emotion Category Output}
\begin{table}[!htb]
\centering
\begin{tabular}{|c|c|c|c|c|c|c|c|c|}
\hline
                                           & \multicolumn{8}{c|}{Accuracy}                                              \\ \hline
                                           & Neutral & Happy & Sad  & Angry & Fear & Surprise & Disgust & Total         \\ \hline
\cellcolor[HTML]{EFEFEF}VGG16+LSTM+1FC          & 0.89    & 0.00  & 0.97 & 0.64  & 0.57 & 0.33     & 0.67    & 0.80          \\ \hline
\cellcolor[HTML]{EFEFEF}VGG16+LSTM+3FC          & 0.91    & 0.00  & 0.99 & 0.50  & 0.41 & 0.00     & 0.23    & 0.75          \\ \hline
\cellcolor[HTML]{EFEFEF}Xcep+LSTM+2FC           & 0.59    & 0.00  & 0.81 & 0.02  & 0.01 & 0.00     & 0.00    & 0.48          \\ \hline
\cellcolor[HTML]{EFEFEF}Xcep+LSTM+3FC           & 0.28    & 0.00  & 0.79 & 0.02  & 0.01 & 0.00     & 0.00    & 0.38          \\ \hline
\cellcolor[HTML]{EFEFEF}VGG19+LSTM+1FC & 0.91    & 0.00  & 0.97 & 0.68  & 0.80 & 0.65     & 0.71    & 0.84 \\ \hline
\cellcolor[HTML]{EFEFEF}VGG19+LSTM+2FC          & 0.94    & 0.01  & 0.98 & 0.61  & 0.78 & 0.14     & 0.73    & 0.82          \\ \hline
\end{tabular}
\caption{Accuracies obtained on training set by various CNN-RNN architectures for combined models}
\label{table:combined_cate_train}
\end{table}

\subsection*{Combined Models Valence \& Arousal Output}
\begin{table}[!htb]
\centering
\begin{tabular}{|c|c|c|c|c|}
\hline
                                             & \multicolumn{2}{c|}{CCC}      & \multicolumn{2}{c|}{MSE}      \\ \hline
                                             & Valence       & Arousal       & Valence       & Arousal       \\ \hline
\cellcolor[HTML]{EFEFEF}VGG16+LSTM+1FC          & 0.78          & 0.68          & 0.11          & 0.05          \\ \hline
\cellcolor[HTML]{EFEFEF}VGG16+LSTM+3FC          & 0.74          & 0.57          & 0.11          & 0.06          \\ \hline
\cellcolor[HTML]{EFEFEF}Xcep+LSTM+2FC           & 0.11          & 0.11          & 0.27          & 0.09          \\ \hline
\cellcolor[HTML]{EFEFEF}Xcep+LSTM+3FC           & 0.07          & 0.07          & 0.27          & 0.09          \\ \hline\cellcolor[HTML]{EFEFEF}VGG19+LSTM+1FC          & 0.83          & 0.71          & 0.09          & 0.05          \\ \hline
\cellcolor[HTML]{EFEFEF}VGG19+LSTM+2FC & 0.80 & 0.67 & 0.10 & 0.05 \\ \hline
\end{tabular}
\caption{CCC and MSE evaluation of valence \& arousal predictions reached by the CNN-RNN architectures in combined models on training set}
\label{table:combined_va_train}
\end{table}

\clearpage

\section{Evaluation of the Models on Testing Set}
\subsection*{Category-only Models}
\begin{table}[!htb]
\centering
\begin{tabular}{|c|c|c|c|c|c|c|c|c|}
\hline
                                           & \multicolumn{8}{c|}{Accuracy}                                              \\ \hline
                                           & Neutral & Happy & Sad  & Angry & Fear & Surprise & Disgust & Total         \\ \hline
\cellcolor[HTML]{EFEFEF}VGG16+LSTM+1FC          & 0.05    & 0.00  & 0.96 & 0.00  & 0.00 & 0.00     & 0.00    & 0.27          \\ \hline
\cellcolor[HTML]{EFEFEF}VGG16+LSTM+2FC          & 0.24    & 0.00  & 0.77 & 0.07  & 0.00 & 0.00     & 0.15    & 0.32           \\ \hline
\cellcolor[HTML]{EFEFEF}Xcep+LSTM+1FC           & 0.07    & 0.00  & 0.52 & 0.00  & 0.00 & 0.00     & 0.00    & 0.17          \\ \hline
\cellcolor[HTML]{EFEFEF}Xcep+LSTM+2FC           & 0.41    & 0.00  & 0.76 & 0.00  & 0.00 & 0.00     & 0.00    & 0.37          \\ \hline
\cellcolor[HTML]{EFEFEF}VGG19+LSTM+1FC & 0.09    & 0.00  & 1.00 & 0.00  & 0.00 & 0.00     & 0.00    & 0.30 \\ \hline
\cellcolor[HTML]{EFEFEF}VGG19+LSTM+2FC          & 0.72    & 0.00  & 0.87 & 0.00  & 0.00 & 0.00     & 0.00    & 0.52          \\ \hline
\end{tabular}
\caption{Accuracies obtained on testing set by various CNN-RNN architectures for categorical-only models}
\label{table:cate_res_test}
\end{table}

\clearpage

\subsection*{Dimension-only Models}
\begin{table}[!htb]
\centering
\begin{tabular}{|c|c|c|c|c|}
\hline
                                    & \multicolumn{2}{c|}{CCC}      & \multicolumn{2}{c|}{MSE}      \\ \hline
                                    & Valence       & Arousal       & Valence       & Arousal       \\ \hline
\cellcolor[HTML]{EFEFEF}VGG16+LSTM+1FC & 0.04          & 0.03         & 0.46          & 0.18          \\ \hline
\cellcolor[HTML]{EFEFEF}VGG16+LSTM+2FC & 0.01         & -0.09          & 0.43        & 0.15         \\ \hline
\cellcolor[HTML]{EFEFEF}Xcep+LSTM+1FC  & 0.02          & 0.06          & 0.29          & 0.10          \\ \hline
\cellcolor[HTML]{EFEFEF}Xcep+LSTM+2FC  & 0.00          & 0.01          & 0.32          & 0.10          \\ \hline
\cellcolor[HTML]{EFEFEF}VGG19+LSTM+1FC & -0.05        & -0.01      & 0.58      & 0.20      \\ \hline
\cellcolor[HTML]{EFEFEF}VGG19+LSTM +2FC & -0.05 & 0.01 & 0.43 & 0.13 \\ \hline
\end{tabular}
\caption{CCC and MSE evaluation of valence \& arousal predictions reached by the CNN-RNN architectures in dimensional-only models on testing set}
\label{table:va_test}
\end{table}

\clearpage

\subsection*{Combined Models Emotion Category Output}
\begin{table}[!htb]
\centering
\begin{tabular}{|c|c|c|c|c|c|c|c|c|}
\hline
                                           & \multicolumn{8}{c|}{Accuracy}                                              \\ \hline
                                           & Neutral & Happy & Sad  & Angry & Fear & Surprise & Disgust & Total         \\ \hline
\cellcolor[HTML]{EFEFEF}VGG16+LSTM+1FC          & 0.18    & 0.00  & 0.62 & 0.01  & 0.00 & 0.00     & 0.06    & 0.24          \\ \hline
\cellcolor[HTML]{EFEFEF}VGG16+LSTM+3FC          & 0.12    & 0.00  & 0.81 & 0.00  & 0.00 & 0.00     & 0.00    & 0.26          \\ \hline
\cellcolor[HTML]{EFEFEF}Xcep+LSTM+2FC           & 0.14    & 0.00  & 0.84 & 0.00  & 0.00 & 0.00     & 0.00    & 0.28          \\ \hline
\cellcolor[HTML]{EFEFEF}Xcep+LSTM+3FC           & 0.15    & 0.00  & 0.72 & 0.00  & 0.00 & 0.00     & 0.00    & 0.25          \\ \hline
\cellcolor[HTML]{EFEFEF}VGG19+LSTM+1FC & 0.13    & 0.00  & 0.86 & 0.02  & 0.00 & 0.00     & 0.06    & 0.29 \\ \hline
\cellcolor[HTML]{EFEFEF}VGG19+LSTM+2FC          & 0.09    & 0.00  & 0.74 & 0.19  & 0.13 & 0.00     & 0.05    & 0.25          \\ \hline
\end{tabular}
\caption{Accuracies obtained on testing set by various CNN-RNN architectures for combined models}
\label{table:combined_cate_test}
\end{table}

\subsection*{Combined Models Valence \& Arousal Output}
\begin{table}[!htb]
\centering
\begin{tabular}{|c|c|c|c|c|}
\hline
                                             & \multicolumn{2}{c|}{CCC}      & \multicolumn{2}{c|}{MSE}      \\ \hline
                                             & Valence       & Arousal       & Valence       & Arousal       \\ \hline
\cellcolor[HTML]{EFEFEF}VGG16+LSTM+1FC          & -0.03          & 0.02          & 0.63          & 0.27          \\ \hline
\cellcolor[HTML]{EFEFEF}VGG16+LSTM+3FC          & -0.06          & -0.15          & 0.55          & 0.20          \\ \hline
\cellcolor[HTML]{EFEFEF}Xcep+LSTM+2FC           & 0.02          & 0.07          & 0.36          & 0.12          \\ \hline
\cellcolor[HTML]{EFEFEF}Xcep+LSTM+3FC           & -0.05          & 0.03          & 0.39          & 0.12          \\ \hline
\cellcolor[HTML]{EFEFEF}VGG19+LSTM+1FC          & -0.05          & 0.05          & 0.54          & 0.26          \\ \hline
\cellcolor[HTML]{EFEFEF}VGG19+LSTM+2FC & -0.12 & -0.02 & 0.59 & 0.18 \\ \hline
\end{tabular}
\caption{CCC and MSE evaluation of valence \& arousal predictions reached by the CNN-RNN architectures in combined models on testing set}
\label{table:combined_va_test}
\end{table}






\end{document}